\documentclass[10pt]{article}
\usepackage[utf8]{inputenc}
\usepackage[T1]{fontenc}
\usepackage{hyperref}
\usepackage{url}
\usepackage{booktabs}
\usepackage{amsfonts}
\usepackage{amsmath}
\usepackage{graphicx}
\usepackage{float}
\usepackage{caption}
\usepackage{subcaption}
\usepackage{multirow}
\usepackage{enumitem}
\usepackage{xcolor}
\usepackage{pifont}
\newcommand{\cmark}{\textcolor{green!45!black}{\ding{51}}}
\newcommand{\xmark}{\textcolor{red!65!black}{\ding{55}}}
\usepackage{placeins}
\usepackage[most]{tcolorbox}

\definecolor{promptframe}{HTML}{2F4B7C}
\definecolor{promptback}{HTML}{F7F9FC}
\colorlet{promptmark}{promptframe!60}
\newcommand{\ph}[1]{{\color{promptframe}\itshape\textlangle #1\textrangle}}
\newtcblisting{promptbox}[1]{%
  enhanced, breakable, listing only,
  colback=promptback, colframe=promptframe, coltitle=white,
  boxrule=0.5pt, arc=1.5pt, left=5pt, right=5pt, top=3pt, bottom=3pt,
  toptitle=1.5pt, bottomtitle=1.5pt,
  fonttitle=\small\sffamily\bfseries, title={#1},
  listing options={basicstyle=\ttfamily\scriptsize, columns=fullflexible,
    keepspaces=true, breaklines=true, showstringspaces=false, upquote=true,
    aboveskip=0pt, belowskip=0pt,
    escapeinside={(*}{*)},
    postbreak=\mbox{\textcolor{promptmark}{$\hookrightarrow$}\space}}}

\newcommand{\passhat}[1]{\text{pass\textasciicircum{}#1}}

\graphicspath{{figures/}}
\newcommand{\MeanSightedPass}{4.15\%}
\newcommand{\MeanBlindPass}{1.82\%}

\newcommand{\PassDropPP}{2.33}
\newcommand{\PassDropCILow}{1.42}
\newcommand{\PassDropCIHigh}{3.41}
\newcommand{\MeanSightedFirst}{26.1\%}
\newcommand{\MeanBlindFirst}{15.1\%}
\newcommand{\MeanSightedInvalid}{2.7\%}
\newcommand{\MeanBlindInvalid}{27.6\%}
\newcommand{\SightedMatchedWin}{13.9\%}
\newcommand{\SightedMismatchWin}{0.8\%}

\newcommand{\BlindMismatchWin}{0.3\%}

\usepackage{floatai-techreport}
\hypersetup{colorlinks=true,breaklinks=true}

\title{Finding the Move Is Not Winning the Game:\\XiangqiBench for Closed-Loop Evaluation of LLM Agents}

\author[1,2]{Yekun Chai}
\author[1,3]{Qiwei Peng}
\author[4]{Haoyi Xiong}
\affiliation[1]{FloatAI}
\affiliation[2]{ETH Zurich}
\affiliation[3]{University of Copenhagen}
\affiliation[4]{Independent Researcher}
\date{\today}
\metadata{Code}{\url{https://github.com/floatai/xiangqibench}}
\reportlogo{\includegraphics[height=0.65cm]{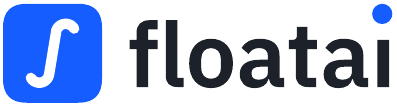}}

\begin{document}
\maketitle
% Shared body for main.tex (ICLR) and main-arxiv.tex (arXiv); the abstract must stay here so
% that floatai-techreport.sty can draw the arXiv title panel.
\begin{abstract}
Static evaluations credit a language model for naming the right move, but an agent must carry
a plan through to a verified outcome while an opponent responds. We introduce
\textbf{XiangqiBench}, an executable benchmark that measures this difference in Chinese chess:
starting from 119 tactical endgames with forced mates supported by engine or checks-only search, an LLM agent must
deliver checkmate against an engine defender. An interactive REPL interface separates real moves,
state queries, and forward simulation, and we record 8,568 multi-turn trajectories from 12
frontier LLMs under two observation protocols. Three signals that look like competence each
overstate closed-loop success.
(i)~\emph{The Conversion Gap}: models play the stored reference first move in 26.1\% of
Sighted trials, yet only 13.9\% of these trials end in a win.
(ii)~\emph{The Consistency Gap}: the leading model reaches 38.7\% pass@3 but only 5.9\%
\passhat{3}, winning all three trials on 7 of the 46 positions it ever wins.
(iii)~\emph{The Simulation Gap}: 32.3\% of accepted simulation calls stop on an illegal move,
and in 49.3\% of comparable cases the real defender replies differently from the line the agent
simulated; self-authored rollouts check legality but cannot anticipate the opponent.
Finding the move is not winning the game: agent evaluations should score closed-loop outcomes
and report reliability alongside coverage.
\end{abstract}

\section{Introduction}
\label{sec:introduction}

\begin{figure}[t]
\centering
\includegraphics[width=0.76\linewidth]{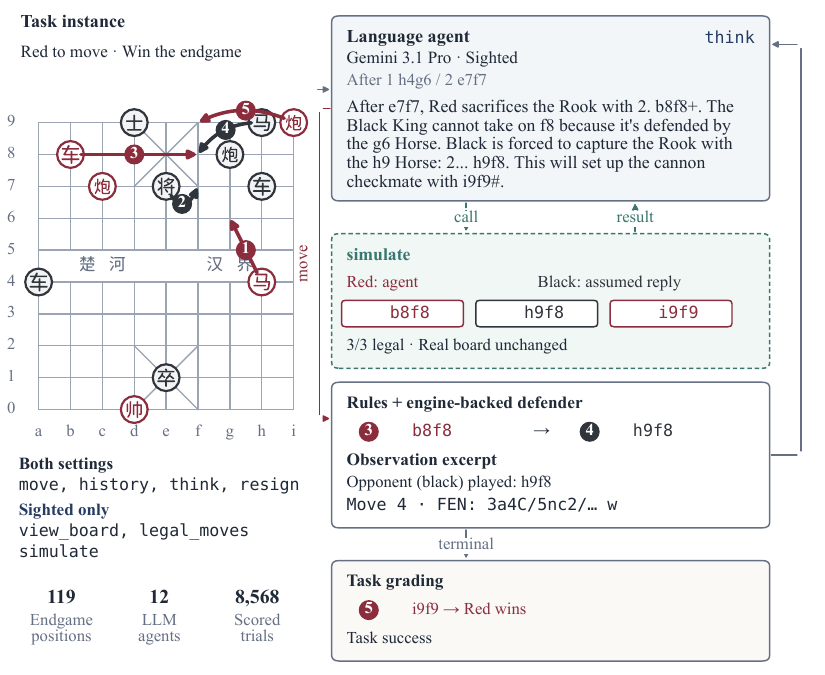}
\caption{\textbf{XiangqiBench at a glance.} \emph{Left:} a task instance with
the real trajectory overlaid on the initial board (numbered plies; deep red for
the agent, charcoal for the defender; curves separate annotations and are not
piece paths), tool availability by setting, and the scale of the evaluation.
\emph{Right:} the second Red decision of
Gemini 3.1 Pro in Sighted trial~3 on case SQYQ-251, after real plies 1--2
(Appendix~\ref{app:gemini-overview}). The complete pre-simulation
\texttt{think} response is shown; \texttt{simulate} then checks three
model-specified plies, including an assumed Black reply, on a copy of the board.
The defender's actual reply informs the next decision, and the terminal Red win
determines task success.}
\label{fig:overview}
\end{figure}

Static evaluations such as puzzle solving and move prediction credit a language model for naming
a good action \citep{chen2025xiangqir1}. Agents are held to a different standard: they must carry a plan through to a
verified outcome while an active adversary responds \citep{llmchess2025,liu2026chessarena}. Success then hinges on
\emph{conversion}: navigating an unfolding contingency tree to a verified goal. Whether static
competence predicts conversion is rarely measured directly, because existing benchmarks often
confound heuristic proposal, mental planning, and closed-loop execution.

Board-game endgames provide a controlled testbed for this decision loop: state transitions
are exact, legality is executable, and outcomes are determined by verifiable terminal states \citep{toshniwal2022chessgpt,ruoss2024amortized,schultz2025mastering}. We study \emph{endgame conversion} in Chinese chess (Xiangqi): starting from a
tactical position in which engine or checks-only search reports a forced mate for Red, an LLM agent plays Red against
an engine defender. Agreement with reference moves receives no score; the agent must sustain a winning
continuation across multiple turns until checkmate \citep{grace2026agentevals}.
XiangqiBench exposes this process through an interactive Read--Evaluate--Print Loop
(REPL) interface (Fig.~\ref{fig:overview}) that separates real moves, state queries of the current
board and legal moves, and a forward-simulation sandbox on a cloned board.
Simulation is not an external search engine: the agent must supply both
its own candidate moves and the hypothesized opponent replies. The tool therefore
shows what the agent expects: does it anticipate the defender's replies, or does it
check lines the real defender does not play?

We evaluate 12 frontier LLMs (Gemini 3.1 Pro \citep{gemini31pro}, GPT-5.5 \citep{openai2026gpt55}, Qwen 3.7 Max \citep{qwen37}, Claude Opus 4.7 \citep{anthropic2026opus47},
DeepSeek V4 Pro/R1/V3.2 \citep{guo2025deepseek,liu2025deepseek,xu2026deepseek}, Seed 2.0 Pro \citep{seed2026seed2}, Kimi K2.5 \citep{team2026kimi}, Mistral Large 3 \citep{mistral3}, MiniMax M2.5 \citep{chen2026minimax}, and \mbox{GPT-OSS} 120B \citep{agarwal2025gpt};
Appendix~\ref{app:models}) under two observation settings:
\emph{Sighted}, which provides board refreshes, legal-move feedback, and
the simulation sandbox; and \emph{Restricted observation}, which withholds board
refreshes, state queries, and simulation to probe state tracking from move history.
Our contributions are:
\begin{itemize}[leftmargin=*,itemsep=0pt,topsep=1pt]
\item \textbf{A benchmark for adversarial conversion:} 119 search-supported tactical endgames, an
executable rules environment with an engine defender, and 8,568 recorded trajectories.
\item \textbf{An instrumented decision loop:} we trace each trial from the first move through
simulation legality and real-move adoption to defender divergence and the terminal outcome.
\item \textbf{Three gaps between apparent and closed-loop competence:} a \emph{Conversion Gap}
(the reference first move rarely leads to a win), a \emph{Consistency Gap} (pass@3 far exceeds
\passhat{3}), and a \emph{Simulation Gap} (self-authored rollouts often break legality and
mispredict the defender); a controlled ablation on the two strongest models suggests that much of
the Restricted-observation drop reflects the setting's prompt and feedback rather than the missing
board alone.
\end{itemize}

\section{Related Work}
\label{sec:related}

\paragraph{Interactive agent evaluation.}
AgentBench, SWE-bench, WebArena, and OSWorld evaluate sequential tool invocation and computer
control \citep{liu2024agentbench,jimenez2024swebench,zhou2024webarena,xie2024osworld}. In game
domains, GTBench, GameBench, GAMEBoT, KORGym, and TextArena provide rule-grounded strategic environments
\citep{duan2024gtbench,costarelli2024gamebench,lin2025gamebot,shi2025korgym,guertler2025textarena}.
XiangqiBench complements them by isolating \emph{adversarial conversion} of search-supported winning
positions against an engine defender, scoring only the terminal outcome while recording intermediate
planning and real-move commitment \citep{grace2026agentevals}.

\paragraph{Game planning, state tracking, and simulation.}
Board games serve as testbeds for neural planning and representation
\citep{toshniwal2022chessgpt,feng2023chessgpt,ruoss2024amortized,schultz2025mastering}.
In chess, ChessArena and LLM Chess evaluate full games against other LLMs, random players, and
engines, and ChessArena adds a blindfold mode that supplies only the move history
\citep{liu2026chessarena,llmchess2025}. LMRL-Gym uses theoretical endgames as a multi-turn RL testbed
\citep{abdulhai2023lmrl}, while Xiangqi-R1 trains a 7B model in multiple stages to improve move
legality and position analysis \citep{chen2025xiangqir1}. We instead study generalist frontier LLMs on
multi-turn endgame conversion and separate the decision loop into scratchpad simulation, real-move
adoption, and resilience against engine counterplay.

\paragraph{Reliability metrics.}
Following \citet{chen2021codex} and \citet{yao2025taubench}, we use unbiased finite-sample estimators
for pass@$k$ (at least one success) and \passhat{k} (all $k$ attempts succeed); contrasting them
reveals whether an agent that can reach a win also reaches it reliably.

\section{XiangqiBench: Task Construction}
\label{sec:benchmark}

\paragraph{Corpus.}
XiangqiBench curates 119 tactical compositions from classical Chinese-chess treatises in which Red can
force checkmate: 116 from \emph{Shi Qing Ya Qu}\footnote{\url{https://www.xqipu.com/canjugupu/1547}}
(case identifiers SQYQ-$n$) and 3 from the classical \emph{Jianghu} collection (JH-$n$).
Each entry specifies an initial position in Forsyth--Edwards Notation (FEN) with Red to move;
historical solution text is withheld from prompts and scoring. During evaluation, an executable rules
adapter verifies every proposed action. Xiangqi legality involves non-local constraints: pieces must
respect palace boundaries and river crossings, cannons require an intervening screen to capture, horses
are blocked by adjacent pieces (``hobbling the horse's leg''), and generals cannot face each other
across an open file.

\paragraph{Search-supported admission.}
Historical treatises can contain transcription errors or overlooked defenses, so source solutions
are not trusted. Instead, each position is screened with \textbf{Pikafish} \citep{pikafish}, a
strong open-source Xiangqi engine derived from Stockfish that combines alpha--beta search with an
NNUE evaluation network. For \textbf{admission}, a single-threaded, time-limited search is run from
each position, and the position is admitted when the engine reports a forced mate for Red. The
reported mate distance and the engine's preferred first move are stored as case metadata; the latter
serves as the reference root move in Section~\ref{sec:results}. Composed positions that the engine
does not confirm may instead be admitted by a checks-only mate search and are labelled separately
(Appendix~\ref{app:reproducibility}). For \textbf{defense}, the same engine family plays Black during
evaluation (Section~\ref{sec:protocol}). Because admission relies on bounded search rather than a
complete proof tree, we describe the positions as \emph{search-supported} rather than formally proved;
all 119 positions were additionally checked manually to confirm that Red can force checkmate.

\begin{figure}[t]
\centering
\includegraphics[width=0.82\linewidth]{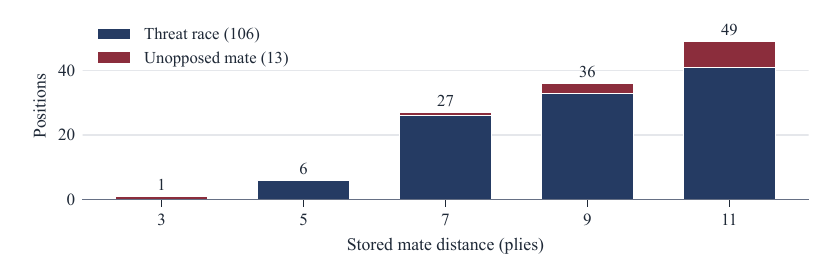}
\caption{\textbf{Stored mate distances of the 119 positions, by category.} Distances are those
reported at admission by Pikafish or the checks-only solver, stacked by category: 106 threat races
(Black would also have a forced mate if it were to move) and 13 unopposed mates (no such counter-mate
found). Distances are admission metadata, not limits on trajectory length during evaluation.}
\label{fig:dataset}
\end{figure}

\paragraph{Why endgames.} Like the chess-endgame task in LMRL-Gym \citep{abdulhai2023lmrl}, we
start from endgames rather than full games: the goal (checkmate) is fixed and reachable, so a failure
to mate is attributable to the agent; opening recall and attrition are removed; and a few precisely
calculated moves decide the game. Unlike LMRL-Gym's few-piece theoretical endgames against a 1200-Elo
engine, ours are composed tactical endgames (median 14 pieces) built on consecutive checks, mating
threats, and pins, and a strong engine defends them, so wins cannot rely on blunders.

\paragraph{Position categories and horizon.} Red can force mate in every position; the two categories
differ only in Black's counterplay. At admission, the same engine or solver is also run with Black to
move. If Black then has a forced mate of its own, the position is a \emph{threat race}: Red must keep
forcing, because a slow move can let Black execute its mate first. Otherwise it is an \emph{unopposed
mate}: no counter-mate was found, so a slow move forfeits tempo but does not by itself hand Black a win.
Stored mate distances range from 3 to 11 plies (median 9; Figure~\ref{fig:dataset}). The 40-ply cap is
more than three times the longest stored distance, leaving room for detours and recovery, and success
already falls steeply within this range (Section~\ref{sec:results}); longer compositions would push
nearly every model to zero, whereas this range still separates the strongest models.

\section{Interaction and Evaluation Protocol}
\label{sec:protocol}

\paragraph{REPL action space.}
The agent plays Red against an engine defender for at most 40 plies. At each turn it receives an
observation $o_t$ (with interaction history $h_t$) and emits free text alongside a command from three
classes (Table~\ref{tab:tools}): transitions $\mathcal{A}_{\text{env}}$, which commit a move in ICCS
coordinates (e.g., \texttt{move h2h9}) and trigger the defender's reply, or resign; read-only queries
$\mathcal{A}_{\text{query}}$, which inspect the board, legal moves, or history without advancing the
game; and hypotheses $\mathcal{A}_{\text{sim}}$, where \texttt{simulate} plays an alternating line of
up to 8 model-supplied plies on a cloned board and reports its legality, and \texttt{think} records
deliberation.

\begin{table}[t]
\centering
\caption{\textbf{REPL action space and observability contract.} Only $\mathcal{A}_{\text{env}}$
mutates the real board.}
\label{tab:tools}
\footnotesize
\setlength{\tabcolsep}{3pt}
\begin{tabular}{@{}lp{0.6\linewidth}cc@{}}
\toprule
Class & Command \& Operational Semantics & Sighted & Restricted \\
\midrule
\multirow{2}{*}{$\mathcal{A}_{\text{env}}$ (Transitions)}
& \texttt{move(from\_to)}: Validate and commit a real move; trigger defender & \cmark & \cmark \\
& \texttt{resign()}: Surrender; terminate trial with defender win & \cmark & \cmark \\
\midrule
\multirow{3}{*}{$\mathcal{A}_{\text{query}}$ (Perception)}
& \texttt{view\_board()}: Render ASCII board, FEN, active color, and piece census & \cmark & \xmark \\
& \texttt{get\_legal\_moves()}: Enumerate legal moves in the current state & \cmark & \xmark \\
& \texttt{get\_history()}: Return the move sequence with piece annotations & \cmark & \cmark \\
\midrule
\multirow{2}{*}{$\mathcal{A}_{\text{sim}}$ (Hypothesis)}
& \texttt{simulate(moves)}: Execute alternating line on cloned state; verify legality & \cmark & \xmark \\
& \texttt{think(text)}: Record uncommitted deliberation & \cmark & \cmark \\
\bottomrule
\end{tabular}
\end{table}

\paragraph{Observation settings.}
Under \textbf{Sighted} observation, the agent receives the updated board, FEN, and legal moves after
every environment step and may use all queries and the simulator; this tests the agent as a system
with external perception. Under \textbf{Restricted observation}, the board and FEN are shown once at
the start; afterwards the agent receives only move coordinates and captures, \texttt{view\_board},
\texttt{get\_legal\_moves}, and \texttt{simulate} are disabled, and it must track the position from the
move history. The two settings also use different system prompts (Appendix~\ref{app:prompt}).

\paragraph{Engine defense and termination.} After each accepted agent move, Black replies with a
fixed-depth (depth-18), single-threaded Pikafish \citep{pikafish} search; environment fallbacks are
described in Appendix~\ref{app:reproducibility}. A trial ends in an agent \textbf{win} if Red
achieves checkmate or leaves Black without legal moves (stalemate is a loss for the stalemated
player). It ends in \textbf{failure} upon Black checkmate, agent resignation, exhausting the 40-ply
budget, exceeding 3 consecutive illegal moves, or threefold repetition. Under official rules,
perpetual checking loses; our replay adjudicates perpetual checks with a simplified long-check rule
(Appendix~\ref{app:rules-check}).

\paragraph{Metrics: exploration vs.\ consistency.} Let $c_i$ denote the successes on case $i$ out of
$n_i = 3$ trials, over $|\mathcal{I}| = 119$ positions. We report \citep{chen2021codex, yao2025taubench}
\begin{equation}
\widehat{\text{pass@}k} = \frac{1}{|\mathcal{I}|} \sum_{i=1}^{|\mathcal{I}|} \left[1 - \frac{\binom{n_i - c_i}{k}}{\binom{n_i}{k}}\right], \qquad
\widehat{\passhat{k}} = \frac{1}{|\mathcal{I}|} \sum_{i=1}^{|\mathcal{I}|} \frac{\binom{c_i}{k}}{\binom{n_i}{k}}.
\end{equation}
pass@$k$ estimates the probability of at least one success in $k$ attempts (exploratory reach), while
\passhat{k} estimates the probability that all $k$ attempts succeed (consistency).

\section{Evaluation Results}
\label{sec:results}

\begin{figure}[t]
\centering
\includegraphics[width=0.82\linewidth]{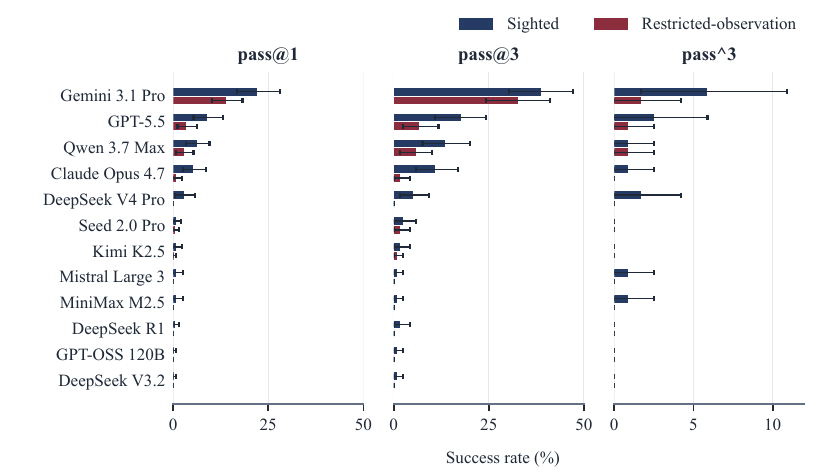}
\caption{\textbf{Multi-turn endgame conversion across frontier LLMs.} Bars report
mean pass@1 (single-trial success), pass@3 (at least one success in 3 attempts), and
\passhat{3} (all 3 attempts successful) across 119 positions (357 trials per model/setting).
Whiskers denote 95\% case-clustered bootstrap intervals.}
\label{fig:results}
\vspace{-1.5em}
\end{figure}

\subsection{The Consistency Gap: Occasional Success vs.\ Reliable Conversion}

Single-trial conversion (pass@1) is low for every model and observation setting
(Figure~\ref{fig:results}). Under Sighted observation, Gemini 3.1 Pro leads at 22.1\% pass@1
(79/357 trials), followed by GPT-5.5 at 9.0\% (32/357) and Qwen 3.7 Max at 6.2\% (22/357).
Under Restricted observation every model declines: Gemini to 14.0\%, GPT-5.5 to 3.4\%, and
Qwen to 2.8\%, with nine models at most 0.8\%.

Repeated attempts separate reaching a win from reproducing it. For Gemini 3.1 Pro under Sighted
observation, pass@3 is 38.7\% (46/119 positions) but \passhat{3} is 5.9\% (7/119): of the 46
positions it wins at least once, it wins all three trials on only 7. For GPT-5.5 the corresponding
counts are 21 and 3 positions (17.6\% and 2.5\%). \passhat{3} is zero in both settings for five
models, and zero under Restricted observation for nine.
Wins are also not spread evenly over positions. If Gemini's 22.1\% per-trial rate held on every
position independently, pass@3 would be 52.7\% and \passhat{3} 1.1\%; the observed 38.7\% and
5.9\% indicate that wins concentrate on a subset of positions. Under Sighted observation, Mistral
Large 3 and MiniMax M2.5 each win one position in all three trials and no other. Within the subset,
outcomes vary from attempt to attempt: on the 46 positions Gemini wins at least once, it wins 57\%
of trials (79/138), and its split of 20, 19, and 7 positions with one, two, and three wins is close
to the 16, 21, and 9 expected from independent trials at that rate. Competence is thus
position-specific, and even where it exists a single run does not reveal it reliably.

\begin{figure}[t]
\centering
\includegraphics[width=0.8\linewidth]{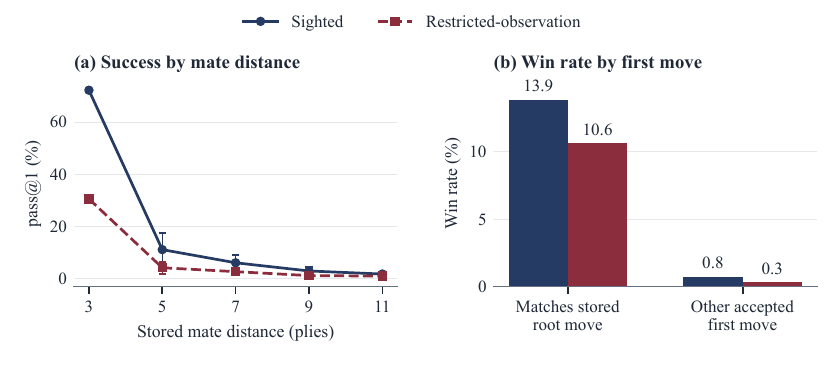}
\caption{\textbf{Deconstructing the Conversion Gap.} (a) Mean pass@1 across the 12 LLMs by stored mate distance,
per observation setting (95\% case-bootstrap intervals; the 3-ply bin contains a single position). (b) Win rate conditioned
on whether the agent's first move matches the stored reference move.}
\label{fig:mechanisms}
\end{figure}

\subsection{The Conversion Gap: Finding the Move vs.\ Winning the Game}

Success falls steeply with the stored mate distance (Figure~\ref{fig:mechanisms}a): Sighted
pass@1, averaged over the 12 LLMs, is 11.1\% at 5 plies, 6.1\% at 7, 2.9\% at 9, and 1.8\% at 11.
The 85 positions at 9 and 11 plies make up 71\% of the benchmark.

To locate where conversion breaks down, we compare the agent's first real move with the engine's
stored reference move (Figure~\ref{fig:mechanisms}b). Across all Sighted trials, the first move
matches the reference in \MeanSightedFirst{} of cases, far above the 3.3\% expected from choosing
uniformly among legal root moves; Gemini 3.1 Pro agrees in 45.1\% of trials and GPT-5.5 in 42.9\%.
A correct first move, however, rarely leads to a win. Trials that begin with the reference move are
won only \SightedMatchedWin{} of the time (155/1,117), so more than 86\% of them are lost later
in the game. Trials that begin with any other accepted move are won in \SightedMismatchWin{}
(Sighted) and \BlindMismatchWin{} (Restricted observation); a different first move is not
necessarily losing, since some positions may admit more than one winning first move. The reference
move is thus close to necessary in practice (155 of the 178 Sighted wins begin with it) but
far from sufficient. The same gap holds within models: Gemini agrees with the reference move in
45.1\% of trials but converts 22.1\%; GPT-5.5 agrees in 42.9\% but converts 9.0\%. We call this the
\textbf{Conversion Gap}: an evaluation that scored only the first move would credit these models
with roughly two to five times the success they achieve over the full game.

\subsection{The Effect of Restricted Observation}

Restricted observation lowers pass@1 for every model. Mean pass@1 drops from \MeanSightedPass{}
in Sighted to \MeanBlindPass{} in Restricted observation, a reduction of \PassDropPP{} percentage
points (95\% case-bootstrap interval [\PassDropCILow{}, \PassDropCIHigh{}]), or more than half.
First-move agreement falls to \MeanBlindFirst{}, and the mean rate of terminal invalid-action or
no-command losses rises from \MeanSightedInvalid{} to \MeanBlindInvalid{} of trials
(Section~\ref{sec:analysis}). Restricted observation, however, also changes the prompt. An ablation
that fixes the prompt and removes the state refreshes and the commands separately
(Appendix~\ref{app:obs-ablation}) shrinks the drop: on positions they solved under Sighted, GPT-5.5
and Gemini without either win 38.1\% and 41.7\% of trials, against 39.7\% and 48.3\% with both
and 9.5\% and 20.0\% under Restricted observation. The drop thus measures the full setting, not the
missing board alone.

\section{Behavioral Analysis}
\label{sec:analysis}

\subsection{How Models Use the REPL Interface}
\label{sec:atomic-capabilities}

\begin{figure}[t]
\centering
\includegraphics[width=0.9\linewidth]{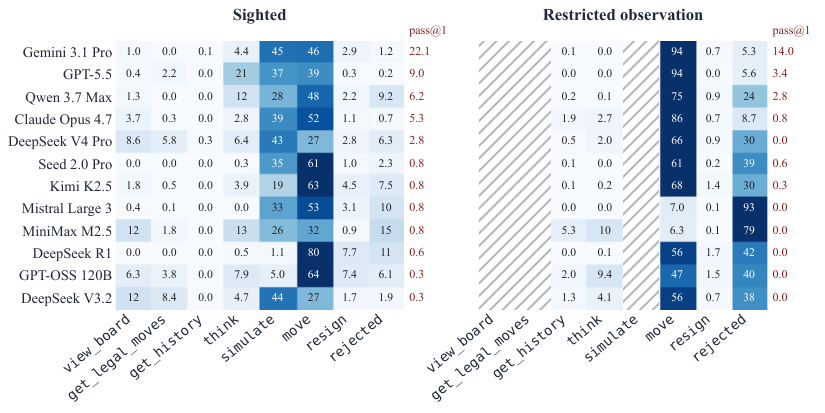}
\caption{\textbf{Action mix by model.} Each cell gives the share (\%) of all agent REPL calls over complete
trajectories spent on each command of Table~\ref{tab:tools}; \texttt{simulate} includes
budget-rejected calls, \texttt{move} counts accepted moves, and \texttt{rejected} counts rejected moves and
the other rejected calls of Figure~\ref{fig:action-phases}. Rows sum to 100\%. Hatched cells are
commands unavailable under Restricted observation. Red numbers give pass@1 (\%).}
\label{fig:behavior-heatmap}
\end{figure}

Figure~\ref{fig:behavior-heatmap} shows how each model distributes its actions across the REPL
commands; three regularities emerge.
\textbf{Simulation volume does not track success.}
Under Sighted observation, \texttt{simulate} accounts for 45\% of Gemini 3.1 Pro's actions, but also for
44\% of DeepSeek V3.2's and 43\% of DeepSeek V4 Pro's, whose pass@1 is 0.3\% and 2.8\%. Seven of the
12 models request simulation in at least 290 of their 357 Sighted trajectories
(Table~\ref{tab:simulation-by-alias}); only DeepSeek R1 (1.1\% of actions) and GPT-OSS 120B (5.0\%)
largely forgo it. Line length does not separate the models either: Qwen 3.7 Max and Mistral Large 3
request the longest lines (mean 4.2 and 3.9 plies), whereas GPT-5.5 simulates a single ply in 71.9\% of
accepted calls (Table~\ref{tab:simulation-length}). Line legality separates the extremes but not the
middle: GPT-5.5 and Gemini obtain at least one fully legal line in 353/356 and 299/332 simulating
trajectories, against 14/213 for Mistral Large 3, yet DeepSeek V3.2 reaches 292/352 and still converts
almost nothing.
\textbf{Explicit state queries are rare and not concentrated in the strongest models.}
Because the Sighted interface already returns the board and legal moves after every step, the two
strongest models almost never call \texttt{view\_board} or \texttt{get\_legal\_moves} (at most 2.2\% of
actions); the heaviest users are DeepSeek V3.2 (12\% and 8.4\%) and DeepSeek V4 Pro (8.6\% and 5.8\%).
\textbf{Rejected actions dominate the weakest models under Restricted observation.}
Without state refreshes, rejected actions remain at 5--9\% of actions for Gemini, GPT-5.5, and Claude
Opus 4.7, rise to 24--42\% for the middle group, and reach 93\% for Mistral Large 3 and 79\% for MiniMax
M2.5 (Figure~\ref{fig:behavior-heatmap}). These two models contribute 2,101 of the 5,850 Restricted rejections
and recover from under 5\% of them, whereas Gemini 3.1 Pro, GPT-5.5, Claude Opus 4.7, and Qwen 3.7 Max
recover from at least 88\% in both settings (Appendix Figure~\ref{fig:error-recovery}; temporal
patterns in Section~\ref{sec:action-phases}).

\subsection{REPL Use Along Trajectories}
\label{sec:trajectories}
\label{sec:action-phases}

\begin{figure}[!htbp]
\centering
\includegraphics[width=0.82\linewidth]{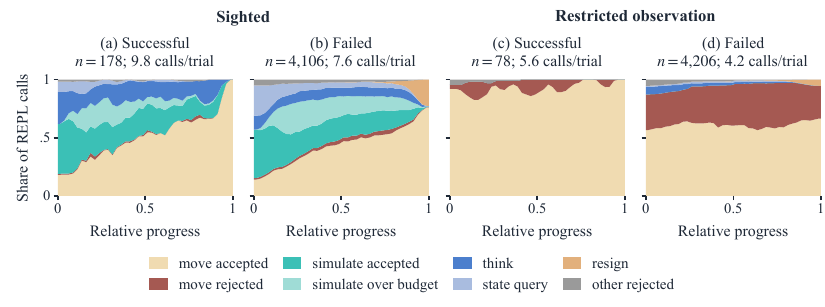}
\caption{\textbf{REPL-call composition along trajectories.} Calls from all 12 LLMs are combined over normalized
trajectory progress $[0, 1]$; each agent call belongs to exactly one category. \emph{State query} counts accepted
read-only perception calls of Table~\ref{tab:tools} (\texttt{view\_board}, \texttt{get\_legal\_moves},
\texttt{get\_history}; only \texttt{get\_history} under Restricted observation); \emph{other rejected} covers
unknown or malformed commands, \texttt{think} calls without reasoning, and over-limit or disabled queries.
Successful Sighted trials devote more calls to \texttt{think} and fewer to over-budget simulation than
failed ones; failed Restricted trials contain a high share of rejected moves throughout.}
\label{fig:action-phases}
\end{figure}

Figure~\ref{fig:action-phases} tracks the composition of REPL calls across relative trial progress $[0, 1]$.
\begin{itemize}[leftmargin=*,itemsep=0pt,topsep=1pt]
\item \textbf{Deliberation versus over-budget simulation (Sighted):} Successful trials spend 14.9\% of their
calls on \texttt{think} (17.4\% in the first quarter of the trajectory), against 7.6\% in failed trials.
Failed trials spend 14.2\% of calls on simulation requests rejected for exceeding the per-turn budget,
against 8.6\% in successes; these requests do not advance the board. In the last quarter of failed
trajectories, resignation rises to 9.3\% of calls.
\item \textbf{Persistent illegality (Restricted):} In failed Restricted trials, rejected moves make up
33.3\% of all calls, against 6.7\% in successful ones. The share is high from the start (29.1\% in
the first quarter, 33.1\% in the last), indicating that state-tracking errors arise early rather than
accumulating only as the position deteriorates.
\end{itemize}
Across the benchmark, the attacker being checkmated or stalemated remains the dominant failure mode
(3,124 Sighted, 2,777 Restricted), while Restricted observation additionally produces 1,178 terminal
invalid-action forfeits (Table~\ref{tab:failure-patterns}).

Per-model breakdowns (Figures~\ref{fig:action-phases-overview}--\ref{fig:action-phases-models-restricted})
show the Sighted contrast most clearly for Gemini 3.1 Pro: its wins spend 14.0\% of calls on \texttt{think}
and 8.7\% on over-budget simulation, its losses 2.4\% and 19.0\%. GPT-5.5 is the exception, interleaving
\texttt{think} and simulation up to its final move whether it wins or loses (\texttt{think} 24.4\% versus
21.2\%). Under Restricted
observation, rejected moves account for 5.2\% and 5.6\% of calls for Gemini 3.1 Pro and GPT-5.5 but
77.6\% for MiniMax M2.5 and 92.9\% for Mistral Large 3. With at most 79 wins per model, these contrasts
are descriptive.

\subsection{The Simulation Gap: Checking Legality Is Not Anticipating the Opponent}
\label{sec:simulation-analysis}

\begin{figure}[t]
\centering
\includegraphics[width=0.9\linewidth]{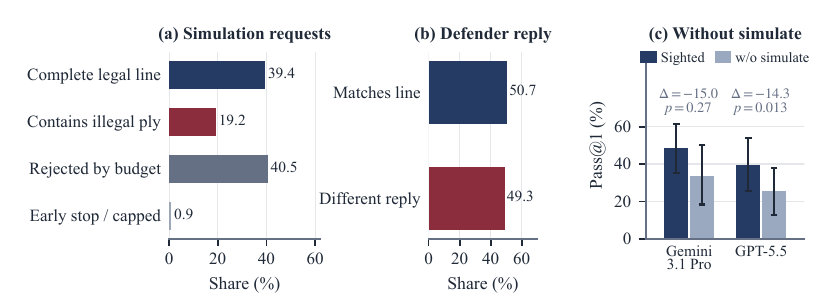}
\caption{\textbf{Simulation checks legality but not the opponent, yet removing it costs wins.}
(a) Outcomes of all 11,342 Sighted simulation requests; ``early stop / capped'' comprises 93 legal
simulated terminals and four legal prefixes truncated at eight plies. (b) For the 1,677 adopted
candidate lines with a legal assumed defender reply, whether the actual defender chose that reply.
(c) Controlled ablation on positions each model solved under Sighted observation, three trials per
position (Appendix~\ref{app:obs-ablation}): Sighted with and without \texttt{simulate}; the ablation
also disables \texttt{view\_board} and \texttt{get\_legal\_moves}, which neither model uses when the
state is refreshed. Bars: pass@1 with 95\% case-bootstrap intervals; $\Delta$: paired difference in
points; $p$: exact sign test over positions.}
\label{fig:simulation-gap}
\vspace{-1.5em}
\end{figure}

A natural hypothesis is that a forward simulation sandbox ($\mathcal{A}_{\text{sim}}$) lets models verify
tactical lines before committing them. The 11,342 simulation requests issued under Sighted observation
reveal two limits on what it verifies (Figure~\ref{fig:simulation-gap}a,b); a controlled ablation shows
that it nonetheless contributes to conversion (Figure~\ref{fig:simulation-gap}c).

\paragraph{Legality falls with line length.} Of 6,749 accepted simulation calls, 4,474 executed the complete candidate sequence legally, 2,178 (32.3\%) stopped on an illegal move, and the remaining 97 ended early at a simulated terminal or the eight-ply cap. Single-ply requests
are almost always legal (96.3\% of 2,592 calls), but full-line legality falls to 47.6\% for 3--4 plies
(1,110/2,332) and 37.4\% for 5--8 plies (357/955). Within their own scratchpads, models move pieces through
palace boundaries, pins, and missing cannon screens.

\paragraph{Assumed opponent replies often differ from the defender's.} Every multi-ply simulated line
contains a defender reply chosen by the model itself. Among the 1,677 cases in which an agent adopted the
first move of a fully legal line on the real board, the actual defender played a different reply in 49.3\%
(826/1,677), so the model's continuation was no longer the line it had checked. These data record
disagreement, not the quality of the assumed reply; the sandbox verifies only the model's
own hypothesis about the opponent, which the defender often does not share.

\paragraph{The sandbox still helps the strongest models.} These limits do not make simulation
useless. We re-ran Gemini 3.1 Pro and GPT-5.5 under Sighted observation with and without
\texttt{simulate}, using a shared prompt and a correct check field, on the 20 and 21 positions each
had solved at least once in the main evaluation (Appendix~\ref{app:obs-ablation}). The ablation also
disables \texttt{view\_board} and \texttt{get\_legal\_moves}, but with the state refreshed after every
step both models almost never call them (at most 0.05 calls per turn, against 0.75 and 0.98
\texttt{simulate} calls). Without \texttt{simulate}, pass@1 falls from 48.3\% to 33.3\% for Gemini
and from 39.7\% to 25.4\% for GPT-5.5 (Figure~\ref{fig:simulation-gap}c). The paired drop is
significant for GPT-5.5 ($p=0.013$) and has the same sign but is not significant for Gemini
($p=0.27$). Simulation thus helps the strongest models find wins even though it verifies only their
own hypothesis about the opponent. Without state refreshes, disabling the same commands does not
lower pass@1 further (Appendix~\ref{app:obs-ablation}).

\begin{figure}[t]
\centering
\includegraphics[width=0.85\linewidth]{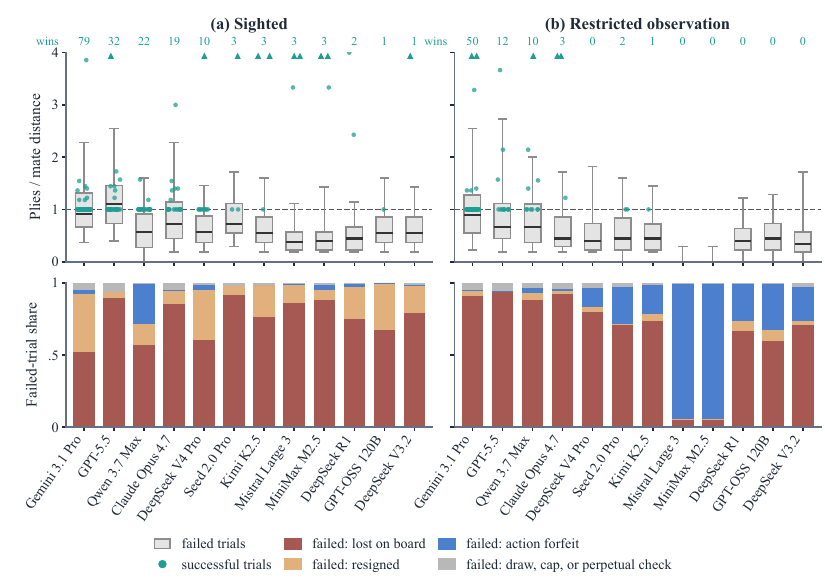}
\caption{\textbf{Failed trajectories end early; successful ones end at the reference length.} Top:
plies played divided by the stored mate distance of the case (dashed line: equal length). Boxes
summarize failed trials (whiskers: 5th--95th percentiles), points show successful trials, and numbers
above the panels give wins out of 357. Triangles mark the 15 wins longer than four times the mate
distance, all in JH-084, where the defender loses by perpetual check. Bottom: how failed trials end;
\emph{lost on board} means the attacker was left without a legal move, and \emph{action forfeit}
covers invalid-action, no-command, and action-limit losses.}
\label{fig:game-length}
\end{figure}

\paragraph{Failing agents are out-raced by the defender.} How games end shows the cost of this gap
(Figure~\ref{fig:game-length}). Wins rarely wander: 142 of 178 Sighted wins and 62 of 78
Restricted-observation wins end exactly at the stored mate distance. Failures end early instead: 77.8\%
of failed Sighted trials and 83.8\% of failed Restricted-observation trials stop before that distance.
Only 80 of the 8,568 trajectories reach ply 40: 76 end at the turn cap (29 Sighted and 47
Restricted-observation), and four end in a repetition draw on that ply. Most failures are lost on the board, and the share
tracks the defender's counter-threat: 81.8\% (Sighted) and 70.4\% (Restricted) of failures in the 106
threat-race positions, against 28.2\% and 29.8\% in the 13 unopposed-mate positions. In the 81 positions
whose defender threatens mate in one, 23.8\% of Sighted and 30.0\% of Restricted-observation trials end at
ply 2: the agent's first move does not stop the threat, and the defender mates at once. Failing agents
are thus not stalled but overtaken by replies they did not anticipate. Stronger models survive longer
even when they lose: across the 12 models, the median relative length of failed trials rises with the
number of wins (Spearman $\rho = 0.71$ Sighted, $0.89$ Restricted observation).

\section{Discussion}
\label{sec:discussion}

The results suggest two explanations for the gap between finding a good move and converting
a won position. Both are interpretations consistent with the data rather than established causes.
\vspace{-2mm}
\paragraph{Amortized policies vs.\ adversarial search.} Converting a won position requires each
move to hold up against the defender's best reply, i.e., an approximation of
$V^*(s_t) = \max_{a_t} \min_{a_{\text{def}}} V^*(s_{t+1})$ over board states $s_t$. A language model instead samples each action
from an amortized policy $P_\theta(a_t \mid o_t, h_t)$ shaped largely by pre-training. Such a policy can
capture familiar motifs, as the 26.1\% first-move agreement shows, but nothing in it checks each
continuation against counterplay. If a model errs with probability $\epsilon_t$ at its $t$-th move,
success on a line of $H$ moves shrinks roughly as $\prod_{t=1}^{H}(1-\epsilon_t)$. This simple picture is
consistent with the steep decline in pass@1 with mate distance and with the gap between Gemini's 45.1\%
first-move agreement and its 22.1\% conversion. It is incomplete: errors are not independent across
moves, and wins concentrate on particular positions.
\vspace{-2mm}
\paragraph{Simulation checks legality, not the opponent.} The agent writes both sides of every
simulated line. A third of accepted simulations stop on an illegal move, and when a legal line began
with the move actually played, the real defender chose a different reply in about half of the cases
(Section~\ref{sec:analysis}). This does not show that the imagined replies were bad, since several
defenses can be equally strong, but it does show that the sandbox cannot supply the opponent's choices.
Agents that rely on self-authored rollouts in adversarial settings still need an external source of
opponent behavior.

\paragraph{Implications for agent evaluation.}
(i)~\textbf{Evaluate closed-loop conversion:} agreement on the first move does not imply the ability to
finish the game, so benchmarks should include play against an active opponent.
(ii)~\textbf{Report reliability alongside coverage:} pass@$k$ and \passhat{k} diverge sharply here and
should be reported together.
(iii)~\textbf{Vary one affordance at a time:} Restricted observation more than halves pass@1, yet a
controlled ablation attributes much of that drop to its prompt and feedback rather than the missing
board, so observation settings should differ only in the affordance under test.

\section{Conclusion}
\label{sec:conclusion}

We introduced \textbf{XiangqiBench}, an executable benchmark in which language agents must convert
search-supported winning Xiangqi endgames into checkmate against an engine defender. Across 8,568
trajectories from 12 frontier LLMs, finding the move is not winning the game: first-move agreement,
pass@3, and self-authored simulation each overstate closed-loop success. Closing these gaps likely
takes more than better single-step proposals; agents must win reliably and anticipate how the
opponent responds. XiangqiBench provides a controlled testbed for measuring both.
\label{sec:main-end}

% \input{sections/limitations}

% \subsubsection*{AI Use Statement}
% We used AI tools to help debug code, proof-read paper, and improve readability. We did not use them to generate data, formulate hypotheses or claims, or design the methodology. All state transitions, legality checks, and game outcomes were computed by the deterministic rules engine and Pikafish, never by a generative AI tool. All AI-assisted code was executed and verified, AI-revised text was rewritten and checked by the authors. We take responsibility for the final content of this work.

% \subsubsection*{Reproducibility Statement.}
% The protocol and metrics are specified in Section~\ref{sec:protocol}; task construction, defender, record
% schema, rules replay, and all known provenance gaps are in Appendix~\ref{app:reproducibility}; models
% and prompts are in Appendices~\ref{app:models} and~\ref{app:prompt-templates}.

\FloatBarrier
\bibliographystyle{iclr2027_conference}
\bibliography{bibliography}
\beginappendix
\makeatletter
\let\xiangqibench@table\table
\let\endxiangqibench@table\endtable
\renewenvironment{table}[1][]{\xiangqibench@table[!htbp]}{\endxiangqibench@table}
\makeatother

\section{Benchmark and Protocol}
\label{app:reproducibility}

\subsection{Positions}

Each case stores an identifier and source, the normalized FEN, the attacking side, the admission method,
one reference root move, the reported mate distance in plies, a position category, and piece tags. Model
outputs play no role in admission. Engine-admitted cases come from a single-threaded, fixed-time Pikafish
search (800\,ms by default) that reports a forced mate; its final preferred move is stored as the reference
move, and every reference move is legal under rules replay. Positions that Pikafish cannot load are
admitted by a checks-only mate search and labelled separately. Search logs, principal variations, and
per-case search settings were not archived, so exact reconstruction would also require the engine build
and network.

The category comes from a second search with Black to move. A \emph{threat race} (106 cases) is a position
in which Black also has a forced mate; an \emph{unopposed mate} (13 cases) is one in which none is found
within the search limit, which is time-limited (300\,ms by default) for engine-admitted cases and
depth-limited for solver-admitted ones.

\subsection{Defender}

On every Black turn the environment sends the current FEN to Pikafish and plays its depth-18,
single-threaded choice. If the engine cannot load the position or returns an error, a rule-based defender
answers instead: it scores each legal Black move with a depth-5 forced-mate search for Red and prefers
moves with no mate found, then the longest mate, then checking moves, breaking ties by move string. The two
solver-admitted cases, JH-084 and JH-327, place a Black pawn on its own back rank, a composed placement
that Pikafish rejects, so the rule-based defender plays almost every Black move in them. These two cases
contribute 39 of the 178 Sighted wins and 17 of the 78 Restricted-observation wins, including 18 of the 20
perpetual-check wins (Appendix~\ref{app:rules-check}). Records log the defender type at initialization
only; per-move provenance is inferred from these engine rejections.

\subsection{Interface, Budgets, and Termination}
\label{app:prompt}

Both settings share the command interface of Table~\ref{tab:tools}; a real move is a coordinate command
such as \texttt{move h2h9}. The Sighted prompt asks for a forcing checkmate. Its first message gives the
board, FEN, and legal moves, and every later message repeats them together with captures and check status.
The Restricted-observation prompt is a generic blindfold prompt that describes the standard opening
position, although the first message shows the actual endgame board and FEN. Later messages report only
the move, capture, check status, and material; board views, legal-move queries, and simulation are
disabled, while \texttt{get\_history} remains available. The prompt mismatch is visible in behavior: 70.5\%
(2,330/3,304) of rejected first-turn moves under Restricted observation are legal only from the standard
opening, against 2.6\% under Sighted observation. Because prompt and observation change together, we treat
the setting comparison as descriptive and isolate observation in Appendix~\ref{app:obs-ablation}. Both
prompts are reproduced in Appendix~\ref{app:prompt-templates}.

Every trial is capped at 40 plies. In the dominant configuration, a turn allows eight actions, one board
view, one simulation of at most eight model-supplied plies, three rejected actions (unknown commands
included), and five consecutive responses without a command. Exhausting any budget forfeits the trial, and
reaching the cap without a win counts as a failure. The Sighted prompt instead states that exceeding the
action limit forces a move; 25 Sighted trials end by action-limit forfeit. Seventy-five Sighted records use
a four-action limit and two a five-rejection limit. Records ending in API errors, timeouts, or exceptions,
and a small number of undetermined or action-budget outcomes, are excluded from scoring.

\subsection{Trajectory Records and Selection}

Each record stores the initial board and FEN, every move with the positions before and after it, the legal
moves at each turn, every prompt, the model's text and commands, provider token usage, environment
feedback, and the terminal result. Simulation results are stored as rendered feedback, from which we
recover per-ply validity. Within each of the 2,856 model--case--setting cells we keep the first three scored
records by start time, giving 8,568 trajectories. This omits eight further records, all failures, in three
Sighted cells of Mistral Large 3 whose retained trials also fail. Requests that exhausted their API retries
leave no record, so the number of attempted requests is unknown.

\subsection{Rules Replay and Repetition}
\label{app:rules-check}

\paragraph{Check flag.} The collection environment set the check flag with the rules library's
\texttt{is\_checking()} predicate after each move. The predicate tests whether the side to move attacks the
opposing general, which never holds after a legal move. All 57,000 \texttt{Check} fields shown to agents,
and every per-ply field in simulation output, therefore read \texttt{No}. The defect leaves board
transitions, legality, and winners unchanged, but it labelled every checkmate as a stalemate and disabled
the collection-time perpetual-check rule.

\paragraph{Replay.} We replay all accepted actions with \texttt{cchess} 1.25.5 after discarding 423
synthetic forfeit entries that repeat the prior move. The replay splits the 6,137 archived stalemates into
6,034 checkmates and 103 stalemates without changing any winner.

\paragraph{Repetition.} With the flag always false, collection scored every threefold repetition as a draw.
We instead apply the released environment's long-check rule: at threefold repetition, a side that gave
check on each of its moves since the first occurrence loses, and any other repetition is a draw. Replaying
the 138 repetition trajectories reproduces every archived position; only terminal labels change. Of the 43
perpetual checks, the 23 by the attacker remain failures (16 Sighted, seven Restricted) and the 20 by the
defender become wins (14 and six), 18 of them in JH-084 under the rule-based defender. Scoring every
repetition as a draw would give 164 Sighted and 72 Restricted-observation wins, mean pass@1 of 3.83\% and
1.68\%, and a setting difference of 2.15 points ([1.38, 2.99]). The rule simplifies official long-check and
long-chase adjudication. Because agents never received a correct check signal, their play may differ from
what it would have been with one; no relabelling can recover that counterfactual.

\subsection{Metrics and Uncertainty}

Pass@1 is the mean of 357 binary outcomes per model and setting. With three trials per case, pass@$k$
uses the estimator of \citet{chen2021codex} and \passhat{k} that of \citet{yao2025taubench}: pass@3 is the
fraction of the 119 cases with at least one win and \passhat{3} the fraction with three. We report
$k \le 3$ only. The plug-in estimate $(c_i/n_i)^k$ \citep{schmid2025passmetrics} is a different
finite-sample estimator and is not used. Intervals for the setting comparison come from a case bootstrap
that resamples the 119 cases with all models, settings, and trials attached (4,000 resamples, seed 2027).

\FloatBarrier
\section{Models}
\label{app:models}

Table~\ref{tab:models} lists the 12 models, each with three scored trials per case in both settings. Names
follow the providers; deployment versions were not independently verified. Kimi K2.6, GLM 5.1, and Seed 2.1
Pro were also attempted but did not complete collection because of credential failures or request
timeouts; they are omitted rather than scored as losses.

\begin{table}[h]
\centering
\caption{\textbf{Evaluated models.} Model IDs are the names under which trajectories are stored;
\emph{Weights} indicates whether weights are publicly released. Each model contributes 357 scored
trajectories per setting.}
\label{tab:models}
\small
\begin{tabular}{llll}
\toprule
Model & Developer & Weights & Model ID \\
\midrule
Gemini 3.1 Pro & Google & Closed & \texttt{gemini-3.1-pro} \\
GPT-5.5 & OpenAI & Closed & \texttt{gpt-5.5} \\
Qwen 3.7 Max & Alibaba & Closed & \texttt{qwen3.7-max} \\
Claude Opus 4.7 & Anthropic & Closed & \texttt{claude-opus-4.7} \\
DeepSeek V4 Pro & DeepSeek & Open & \texttt{deepseek-v4-pro} \\
Seed 2.0 Pro & ByteDance & Closed & \texttt{seed-2.0-pro} \\
Kimi K2.5 & Moonshot AI & Open & \texttt{kimi-k2.5} \\
Mistral Large 3 & Mistral AI & Open & \texttt{mistral-large-3} \\
MiniMax M2.5 & MiniMax & Open & \texttt{minimax-m2.5} \\
DeepSeek R1 & DeepSeek & Open & \texttt{deepseek-r1} \\
GPT-OSS 120B & OpenAI & Open & \texttt{gpt-oss-120b} \\
DeepSeek V3.2 & DeepSeek & Open & \texttt{deepseek-v3.2} \\
\bottomrule
\end{tabular}
\end{table}

\FloatBarrier
\section{Prompt Templates}
\label{app:prompt-templates}

This appendix reproduces the agent's prompts from the trajectory records; the release includes the script
that generates it. The Restricted-observation prompt is identical across all 4,284 trajectories. The
Sighted prompt shown is used in 4,209; the other 75 state a four-action rather than an eight-action limit.
Per-turn messages appear as templates with \ph{placeholders} for content that changes between turns. For
typesetting we omit the Chinese glosses that accompany English terms, replace the board diagram (drawn
with Chinese glyphs) with a placeholder, and render Unicode punctuation in ASCII; the wording is otherwise
unchanged.

The prompts depart from the evaluated configuration in four places. The Sighted prompt says that
exceeding the action limit forces a move, whereas the configuration forfeits the trial. Both prompts call
the 40-ply limit a draw, which is scored as a failure to convert. The Restricted-observation prompt says the
board is never shown, although the first message shows it once, and states that three consecutive checks
lose, a rule that never fired and is not used in scoring. Finally, every \texttt{Check} field reads
\texttt{No} because of the defect in Appendix~\ref{app:rules-check}.

\subsection{System Prompts}

% Generated by scripts/make_prompt_appendix.py; do not edit by hand.
\begin{promptbox}{Sighted system prompt}
You are solving a Chinese Chess (Xiangqi) ENDGAME.

This is NOT the opening -- it is a sharp endgame position and it is YOUR move.
You are the side to move. Your task is to find the precise forcing sequence
that CHECKMATES the opponent. Play exactly: in these positions a single
slack, non-forcing move can let the opponent escape or launch a decisive
counterattack. There is no draw available -- play to win by checkmate.

================================================================================
1. GAME RULES
================================================================================

Board: 9 columns (a-i) x 10 rows (0-9), divided by a River.
Red occupies rows 0-4 (bottom), Black occupies rows 5-9 (top).
Each side has a Palace -- the 3x3 area around the King.

Pieces and Movement:
  K/k  King     -- One step orthogonally, confined to the Palace. The two
                   Kings may NEVER face each other on the same column with
                   no piece between them (Flying General).
  A/a  Advisor  -- One step diagonally, confined to the Palace.
  B/b  Elephant -- Two steps diagonally. Cannot cross the River.
                   Blocked if the midpoint is occupied.
  N/n  Horse    -- "L" shape. Blocked if the first
                   orthogonal square is occupied.
  R/r  Rook     -- Any distance orthogonally; cannot jump.
  C/c  Cannon   -- Moves like a Rook; to CAPTURE it must jump exactly one
                   piece (the "cannon platform").
  P/p  Pawn     -- Forward one step; after crossing the River also sideways.
                   Never backward.

Win / Loss:
  - Checkmate: opponent's King is in check with no legal escape -> you WIN.
  - Stalemate: opponent has no legal move (even if not in check) -> opponent
    LOSES (unlike Western chess).
  - Move limit: if the game runs past 40 half-moves it is scored as a
    draw (i.e. you failed to convert) -- so do not dawdle.

================================================================================
2. MOVE FORMAT (ICCS)
================================================================================

Each move is exactly 4 characters: <from_col><from_row><to_col><to_row>
  Columns: a b c d e f g h i  (left -> right)
  Rows:    0 1 2 3 4 5 6 7 8 9 (bottom -> top, Red's back rank = 0)
Examples: h2e2 (cannon h2->e2), b0c2 (horse b0->c2), a0a1 (rook a0->a1).

================================================================================
3. BOARD STATE -- WHAT TO TRUST
================================================================================

You get THREE views; trust them in this order:
  1. PIECE POSITIONS census -- coordinate-keyed ground truth of every piece. This
     is authoritative; do NOT rely on memory (a common failure is misreading a
     glyph, e.g. confusing (*\ph{King glyph}*) with (*\ph{Elephant glyph}*)).
  2. FEN string -- compact full-board encoding (ranks listed row 9 -> row 0;
     N=Horse, B=Elephant).
  3. ASCII board -- Chinese glyphs; useful for intuition but easy to misread.
Reconcile your mental model with the census/FEN before every move.

================================================================================
4. AVAILABLE TOOLS
================================================================================

Each turn you MUST eventually call `move` (or `resign`) to end your turn. Before
that you SHOULD `think`, and you MAY use auxiliary tools:

  think(reasoning)      -- **ALWAYS call this first.** Record your analysis of the
                           mating idea, candidate forcing moves, and the lines you
                           calculated. Does not end your turn.
  move(from_to)         -- Play a move. Ends your turn.
  resign()              -- Give up (you lose). Only if the position is truly lost.
  view_board()          -- Show the board with FEN (max 1/turn).
  simulate(moves)       -- Test a hypothetical line WITHOUT affecting the real
                           game. YOU supply both sides: [your_move, assumed_reply,
                           your_move, ...]. Each entry is one half-move. Returns per
                           ply: what moved/captured, check, FEN, and the number of
                           legal replies. Max 8 half-moves,
                           1/turn. Use it to verify a forced mate
                           before committing.
  get_legal_moves()     -- List all legal moves in the current position.
  get_history()         -- Show the full move history.

Recommended flow: think -> (simulate to verify the mate) -> move

After each `move` the environment returns objective facts only (it never tells
you who is winning -- assessing the position is YOUR job): what moved/captured,
whether it is check, material balance, the updated FEN, and the legal moves for
the next position.

================================================================================
5. CONSTRAINTS
================================================================================

  - You can ONLY act via the provided commands/tools; plain text is rejected.
  - You MUST call `think` BEFORE `move`.
  - A turn ends only on `move` or `resign`. Auxiliary tools do not end the turn.
  - If you exceed 8 actions in a turn without moving, a move is
    forced on your behalf.
  - Illegal moves are rejected with the legal-move list; choose a legal move.
  - Keep the attack forcing. Quiet moves usually throw away the win.


================================================================================
INTERACTION PROTOCOL (read carefully)
================================================================================

You control the game through a command-line interface, like a terminal.
This is the ONLY way you may act. Ignore any mention elsewhere of JSON tool
calls or `function(args)` syntax -- in this game you act by typing commands.

On each of your turns:
  1. Think out loud in plain text (your private reasoning -- the opponent never
     sees it).
  2. Then issue EXACTLY ONE command inside a fenced code block:

```bash
move h2e2
```

The environment runs your command and replies with terminal output (prefixed
with `$ <command>`), then it is your turn again until you end it with a move.

Available commands (one per block):
  move <from><to>            Play a move in ICCS format (e.g. `move h2e2`). ENDS your turn.
  simulate <m1> <m2> ...     Forward-test a hypothetical line you supply for BOTH
                             sides: [your move, assumed opponent move, ...]
                             (e.g. `simulate h2e2 h9g7 b0c2`).
  view_board                 Show the board, FEN, and the ground-truth piece list.
  legal_moves                List all legal moves right now.
  think <reasoning>          Record private reasoning (does NOT end your turn).
  history                    Show the move history.
  resign                     Forfeit the game. ENDS your turn.

Rules:
  - Put your command in a ```bash``` block. Only the LAST block is executed.
  - One command per turn-step. To look before you leap, send `view_board` or
    `simulate ...` first, read the output, then send `move ...`.
  - If you emit no command block, you will be asked again.
  - Illegal or unknown commands are rejected with the list of legal moves; keep
    trying -- you do NOT lose your turn for one mistake, but repeated invalid
    commands in a single turn will be resolved against you.
\end{promptbox}

% Generated by scripts/make_prompt_appendix.py; do not edit by hand.
\begin{promptbox}{Restricted-observation system prompt}
You are playing Chinese Chess (Xiangqi) in BLINDFOLD mode.
You interact with the game environment by typing commands in a terminal (see INTERACTION PROTOCOL below).
You CANNOT see the board, FEN, or piece census at any point. You must track the
entire position in your own reasoning, using only the move history and the
per-move diffs you are given. Your memory and mental board-tracking ARE the test.

================================================================================
1. GAME RULES
================================================================================

Board: 9 columns (a-i) x 10 rows (0-9), divided by a River.
Red occupies rows 0-4 (bottom), Black occupies rows 5-9 (top).
Each side has a Palace -- the 3x3 area around the King.

Pieces and Movement:
  King     -- One step orthogonally, confined to the Palace. The two
              Kings may NEVER face on the same file with nothing between
              them (Flying General).
  Advisor  -- One step diagonally, confined to the Palace.
  Elephant -- Exactly two steps diagonally. CANNOT cross the
              River. Blocked if the midpoint is occupied.
  Horse    -- "L" shape. Blocked if the first orthogonal step is
              occupied.
  Rook     -- Any distance orthogonally. Cannot jump.
  Cannon   -- Moves like a Rook; to CAPTURE it must jump exactly one
              piece (the platform).
  Pawn     -- Before the River: one step forward only. After crossing:
              one step forward OR sideways. Never backward.

Win / Loss / Draw:
  - Checkmate: opponent's King in check with no legal escape -> you win.
  - Stalemate: opponent has no legal move but is NOT in check -> opponent
    LOSES (unlike Western chess).
  - Perpetual check: checking 3+ times in a row -> the checking side loses.
  - Threefold repetition -> draw (or the aggressor loses).
  - Move limit: exceeding 40 half-moves -> draw.

================================================================================
2. COORDINATES & INITIAL POSITION
================================================================================

Columns a-i (left->right), rows 0-9 (bottom->top). Red's back rank is row 0,
Black's back rank is row 9. Moves are 4-char ICCS, e.g. 'h2e2'.

Red:   King e0 | Advisors d0 f0 | Elephants c0 g0 | Horses b0 h0 | Rooks a0 i0
       Cannons b2 h2 | Pawns a3 c3 e3 g3 i3
Black: King e9 | Advisors d9 f9 | Elephants c9 g9 | Horses b9 h9 | Rooks a9 i9
       Cannons b7 h7 | Pawns a6 c6 e6 g6 i6

================================================================================
3. CONSTRAINTS
================================================================================

- You have NO board viewer, NO simulator, and NO legal-move list. The ONLY
  game-changing action is `move`. (You may also `think` to record reasoning.)
- After each move you get only a diff: what moved/was captured, check, material.
  After the opponent moves you get the same diff for their move -- never a board.
- If you make 3 illegal moves in a single turn, you FORFEIT.
- You cannot see your opponent's reasoning -- only their moves.


================================================================================
INTERACTION PROTOCOL (read carefully)
================================================================================

You control the game through a command-line interface, like a terminal.
This is the ONLY way you may act. Ignore any mention elsewhere of JSON tool
calls or `function(args)` syntax -- in this game you act by typing commands.

On each of your turns:
  1. Think out loud in plain text (your private reasoning -- the opponent never
     sees it).
  2. Then issue EXACTLY ONE command inside a fenced code block:

```bash
move h2e2
```

The environment runs your command and replies with terminal output (prefixed
with `$ <command>`), then it is your turn again until you end it with a move.

BLINDFOLD: there is no board viewer, simulator, or legal-move list. Track the position from the move history and per-move diffs alone. Repeated illegal moves in one turn forfeit the game.

Available commands (one per block):
  move <from><to>            Play a move in ICCS format (e.g. `move h2e2`). ENDS your turn.
  think <reasoning>          Record private reasoning (does NOT end your turn).
  history                    Show the move history.
  resign                     Forfeit the game. ENDS your turn.

Rules:
  - Put your command in a ```bash``` block. Only the LAST block is executed.
  - If you emit no command block, you will be asked again.
  - Illegal or unknown commands are rejected; repeated invalid commands in a single turn will be resolved against you.
\end{promptbox}

\subsection{Per-Turn Messages}

In both settings the first user message shows the initial position. Sighted feedback then repeats the
full state after every ply; Restricted-observation feedback reports only the move.

\begin{promptbox}{Sighted \textbar{} first user message}
Game started. You are playing as RED.

Initial board:

(*\ph{board diagram with file letters a--i and rank numbers 0--9}*)

Current turn: Red (Uppercase)
Move format: (*\ph{gloss of the ICCS format, e.g.\ h2e2 moves the piece on h2 to e2}*)

FEN: (*\ph{FEN}*) w

Legal moves ((*\ph{n}*)): (*\ph{comma-separated ICCS moves}*)

Type one command inside a ```bash``` block to act (e.g. `move h2e2`). You may `view_board`, `simulate ...`, or `legal_moves` first, then `move` to end your turn.
\end{promptbox}

\begin{promptbox}{Sighted \textbar{} after the agent's move, then after the defender's reply}
Move executed: (*\ph{move}*)
Result: Red (*\ph{piece}*) (*\ph{from}*)->(*\ph{to}*) (*\ph{no capture | captures piece}*)
Capture: you captured (*\ph{piece}*).                      (*\ph{only on a capture}*)
Check: (*\ph{Yes -- you are giving check! | No}*)
Material -- Red: R(*\ph{n}*) N(*\ph{n}*) B(*\ph{n}*) A(*\ph{n}*) C(*\ph{n}*) P(*\ph{n}*) | Black: (*\ph{same fields}*) | Balance: (*\ph{side}*) (*\ph{value}*) (nominal)
Board after your move:

(*\ph{board diagram}*)

Current turn: Black (Lowercase)
Move format: (*\ph{gloss of the ICCS format}*)
FEN: (*\ph{FEN}*) b
Legal moves ((*\ph{n}*)): (*\ph{Black's legal moves}*)

(*\textcolor{promptmark}{\rule[0.5ex]{0.97\linewidth}{0.3pt}}*)
Opponent (black) played: (*\ph{move}*) -- Black (*\ph{piece}*) (*\ph{from}*)->(*\ph{to}*) (*\ph{no capture | captures piece}*)
Check: (*\ph{Yes | No}*)
You LOST your (*\ph{piece}*) on this move.               (*\ph{only on a capture}*)
Move (*\ph{k}*) -- FEN: (*\ph{FEN}*) w

Current board:

(*\ph{board diagram}*)

Current turn: Red (Uppercase)
Move format: (*\ph{gloss of the ICCS format}*)

Piece positions (ground truth -- trust this over your memory):
  Red : (*\ph{piece-type list with coordinates, e.g.\ King e0 | Rook i5 | Cannon g9,h5}*)
  Black: (*\ph{piece-type list with coordinates}*)

Legal moves ((*\ph{n}*)): (*\ph{comma-separated ICCS moves}*)

It is now your turn. Call a tool to proceed.
\end{promptbox}

\begin{promptbox}{Restricted observation \textbar{} first user message}
Game started. You are playing as RED.

Initial board:

(*\ph{board diagram with file letters a--i and rank numbers 0--9}*)

Current turn: Red (Uppercase)
Move format: (*\ph{gloss of the ICCS format, e.g.\ h2e2 moves the piece on h2 to e2}*)

FEN: (*\ph{FEN}*) w

BLINDFOLD: this is the ONLY time you are shown the board. From now on you get only per-move diffs -- no board, FEN, or legal moves. Track the position yourself.

Type one command inside a ```bash``` block to act (e.g. `move h2e2`). You may `think ...` first, then `move` to end your turn. There is no board viewer or legal-move list.
\end{promptbox}

\begin{promptbox}{Restricted observation \textbar{} after the agent's move, then after the defender's reply}
Move executed: (*\ph{move}*)
Result: Red (*\ph{piece}*) (*\ph{from}*)->(*\ph{to}*) (*\ph{no capture | captures piece}*)
Capture: you captured (*\ph{piece}*).                      (*\ph{only on a capture}*)
Check: (*\ph{Yes -- you are giving check! | No}*)
Material -- Red: R(*\ph{n}*) N(*\ph{n}*) B(*\ph{n}*) A(*\ph{n}*) C(*\ph{n}*) P(*\ph{n}*) | Black: (*\ph{same fields}*) | Balance: (*\ph{side}*) (*\ph{value}*) (nominal)

(*\textcolor{promptmark}{\rule[0.5ex]{0.97\linewidth}{0.3pt}}*)
Opponent (black) played: (*\ph{move}*) -- Black (*\ph{piece}*) (*\ph{from}*)->(*\ph{to}*) (*\ph{no capture | captures piece}*)
Check: (*\ph{Yes -- you are in check! | No}*)
(*\ph{No capture. | You LOST your piece on this move.}*)
It is now your turn.
\end{promptbox}

\FloatBarrier
\section{Additional Results}
\label{app:results}

Table~\ref{tab:results} gives pass@1, pass@3, and \passhat{3} for every model and setting, and
Table~\ref{tab:pass-k-full} the full pass@$k$ and \passhat{k} for $k = 1, 2, 3$.

\begin{table}[t]
\centering
\caption{\textbf{Three-trial success and consistency (\%).}
Pass@3 is the fraction of cases with at least one success; \passhat{3} is the
fraction for which all three trials succeed. S/R denote Sighted/Restricted observation.}
\label{tab:results}
\footnotesize
\setlength{\tabcolsep}{1.8pt}
\begin{tabular}{lrrrrrr}
\toprule
& \multicolumn{2}{c}{Pass@1} & \multicolumn{2}{c}{Pass@3}
& \multicolumn{2}{c}{\passhat{3}} \\
\cmidrule(lr){2-3}\cmidrule(lr){4-5}\cmidrule(lr){6-7}
Model & S & R & S & R & S & R \\
\midrule
Gemini 3.1 Pro & 22.1 & 14.0 & 38.7 & 32.8 & 5.9 & 1.7 \\
GPT-5.5 & 9.0 & 3.4 & 17.6 & 6.7 & 2.5 & 0.8 \\
Qwen 3.7 Max & 6.2 & 2.8 & 13.4 & 5.9 & 0.8 & 0.8 \\
Claude Opus 4.7 & 5.3 & 0.8 & 10.9 & 1.7 & 0.8 & 0.0 \\
DeepSeek V4 Pro & 2.8 & 0.0 & 5.0 & 0.0 & 1.7 & 0.0 \\
Seed 2.0 Pro & 0.8 & 0.6 & 2.5 & 1.7 & 0.0 & 0.0 \\
Kimi K2.5 & 0.8 & 0.3 & 1.7 & 0.8 & 0.0 & 0.0 \\
Mistral Large 3 & 0.8 & 0.0 & 0.8 & 0.0 & 0.8 & 0.0 \\
MiniMax M2.5 & 0.8 & 0.0 & 0.8 & 0.0 & 0.8 & 0.0 \\
DeepSeek R1 & 0.6 & 0.0 & 1.7 & 0.0 & 0.0 & 0.0 \\
GPT-OSS 120B & 0.3 & 0.0 & 0.8 & 0.0 & 0.0 & 0.0 \\
DeepSeek V3.2 & 0.3 & 0.0 & 0.8 & 0.0 & 0.0 & 0.0 \\
\bottomrule
\end{tabular}
\end{table}

\begin{table}[t]
\centering
\caption{\textbf{Full finite-sample consistency metrics (\%).}
For each case with $n=3$ trials and $c$ successes,
Pass@$k=1-\binom{3-c}{k}/\binom{3}{k}$ and
\passhat{k} $=\binom{c}{k}/\binom{3}{k}$; entries average the case values.
No metric with $k>3$ is estimated.}
\label{tab:pass-k-full}
\footnotesize
\setlength{\tabcolsep}{2.2pt}
\begin{tabular}{llrrrrrr}
\toprule
& & \multicolumn{3}{c}{Pass@$k$} & \multicolumn{3}{c}{\passhat{k}} \\
\cmidrule(lr){3-5}\cmidrule(lr){6-8}
Model & Prot. & $k=1$ & $k=2$ & $k=3$ & $k=1$ & $k=2$ & $k=3$ \\
\midrule
Gemini 3.1 Pro & S & 22.1 & 33.1 & 38.7 & 22.1 & 11.2 & 5.9 \\
Gemini 3.1 Pro & R & 14.0 & 24.4 & 32.8 & 14.0 & 3.6 & 1.7 \\
GPT-5.5 & S & 9.0 & 14.0 & 17.6 & 9.0 & 3.9 & 2.5 \\
GPT-5.5 & R & 3.4 & 5.3 & 6.7 & 3.4 & 1.4 & 0.8 \\
Qwen 3.7 Max & S & 6.2 & 10.4 & 13.4 & 6.2 & 2.0 & 0.8 \\
Qwen 3.7 Max & R & 2.8 & 4.5 & 5.9 & 2.8 & 1.1 & 0.8 \\
Claude Opus 4.7 & S & 5.3 & 8.7 & 10.9 & 5.3 & 2.0 & 0.8 \\
Claude Opus 4.7 & R & 0.8 & 1.4 & 1.7 & 0.8 & 0.3 & 0.0 \\
DeepSeek V4 Pro & S & 2.8 & 3.9 & 5.0 & 2.8 & 1.7 & 1.7 \\
DeepSeek V4 Pro & R & 0.0 & 0.0 & 0.0 & 0.0 & 0.0 & 0.0 \\
Seed 2.0 Pro & S & 0.8 & 1.7 & 2.5 & 0.8 & 0.0 & 0.0 \\
Seed 2.0 Pro & R & 0.6 & 1.1 & 1.7 & 0.6 & 0.0 & 0.0 \\
Kimi K2.5 & S & 0.8 & 1.4 & 1.7 & 0.8 & 0.3 & 0.0 \\
Kimi K2.5 & R & 0.3 & 0.6 & 0.8 & 0.3 & 0.0 & 0.0 \\
Mistral Large 3 & S & 0.8 & 0.8 & 0.8 & 0.8 & 0.8 & 0.8 \\
Mistral Large 3 & R & 0.0 & 0.0 & 0.0 & 0.0 & 0.0 & 0.0 \\
MiniMax M2.5 & S & 0.8 & 0.8 & 0.8 & 0.8 & 0.8 & 0.8 \\
MiniMax M2.5 & R & 0.0 & 0.0 & 0.0 & 0.0 & 0.0 & 0.0 \\
DeepSeek R1 & S & 0.6 & 1.1 & 1.7 & 0.6 & 0.0 & 0.0 \\
DeepSeek R1 & R & 0.0 & 0.0 & 0.0 & 0.0 & 0.0 & 0.0 \\
GPT-OSS 120B & S & 0.3 & 0.6 & 0.8 & 0.3 & 0.0 & 0.0 \\
GPT-OSS 120B & R & 0.0 & 0.0 & 0.0 & 0.0 & 0.0 & 0.0 \\
DeepSeek V3.2 & S & 0.3 & 0.6 & 0.8 & 0.3 & 0.0 & 0.0 \\
DeepSeek V3.2 & R & 0.0 & 0.0 & 0.0 & 0.0 & 0.0 & 0.0 \\
\bottomrule
\end{tabular}
\end{table}

\subsection{Invalid-Action Losses}
\label{app:failure}

Table~\ref{tab:invalid} reports the share of trials that end in an invalid action or a missing command.
These forfeits do not track strength in either direction: Qwen 3.7 Max wins more Sighted trials than most
models yet loses 25.5\% of them this way, whereas several models rarely forfeit and also rarely win.

\begin{table}[t]
\centering
\caption{Invalid/no-command loss rate (\%) by observation setting. Check feedback is affected as described in Appendix~\ref{app:rules-check}. Records classified as aborted or unfinished are excluded before selecting
three scored records per cell;
each denominator is 357 scored trials.}
\label{tab:invalid}
\scriptsize
\setlength{\tabcolsep}{5pt}
\begin{tabular}{lrr}
\toprule
Model & Sighted & Restricted-observation \\
\midrule
Gemini 3.1 Pro & 2.2 & 0.3 \\
GPT-5.5 & 0.0 & 0.6 \\
Qwen 3.7 Max & 25.5 & 3.4 \\
Claude Opus 4.7 & 0.0 & 1.1 \\
DeepSeek V4 Pro & 0.0 & 13.2 \\
Seed 2.0 Pro & 0.0 & 25.2 \\
Kimi K2.5 & 0.0 & 19.6 \\
Mistral Large 3 & 1.1 & 93.8 \\
MiniMax M2.5 & 0.8 & 93.8 \\
DeepSeek R1 & 2.2 & 25.2 \\
GPT-OSS 120B & 0.8 & 31.9 \\
DeepSeek V3.2 & 0.0 & 23.5 \\
\bottomrule
\end{tabular}
\end{table}

\FloatBarrier
\subsection{Repeated Trials}
\label{app:trial-profiles}

\begin{figure}[!htbp]
\centering
\includegraphics[width=\linewidth]{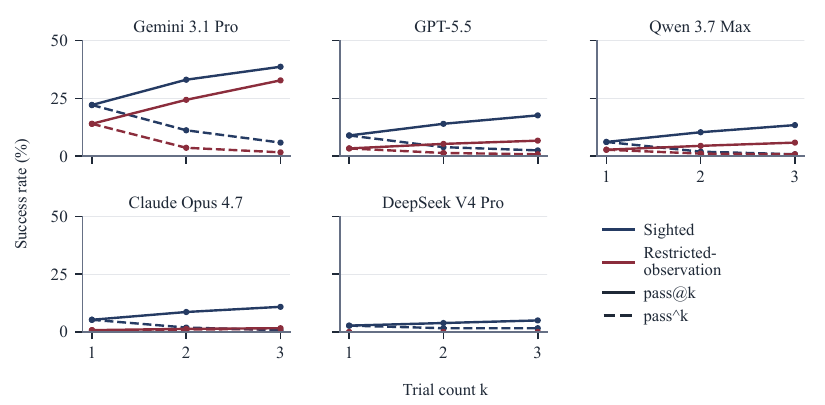}
\caption{\textbf{Reach and reliability diverge with $k$.} Pass@$k$ (solid) and \passhat{k} (dashed) for
$k = 1, 2, 3$, for the five models with the highest Sighted pass@1, computed from the same three trials per
case.}
\label{fig:attempt-budget}
\end{figure}

For Gemini 3.1 Pro under Sighted observation, pass@$k$ rises from 22.1\% to 33.1\% and 38.7\% for
$k = 1, 2, 3$, while \passhat{3} is 5.9\% (Figure~\ref{fig:attempt-budget}). Three trials do not support
extrapolation to larger budgets.

\subsection{Evaluation Set Size}
\label{app:sample-size}

\begin{figure}[!htbp]
\centering
\includegraphics[width=\linewidth]{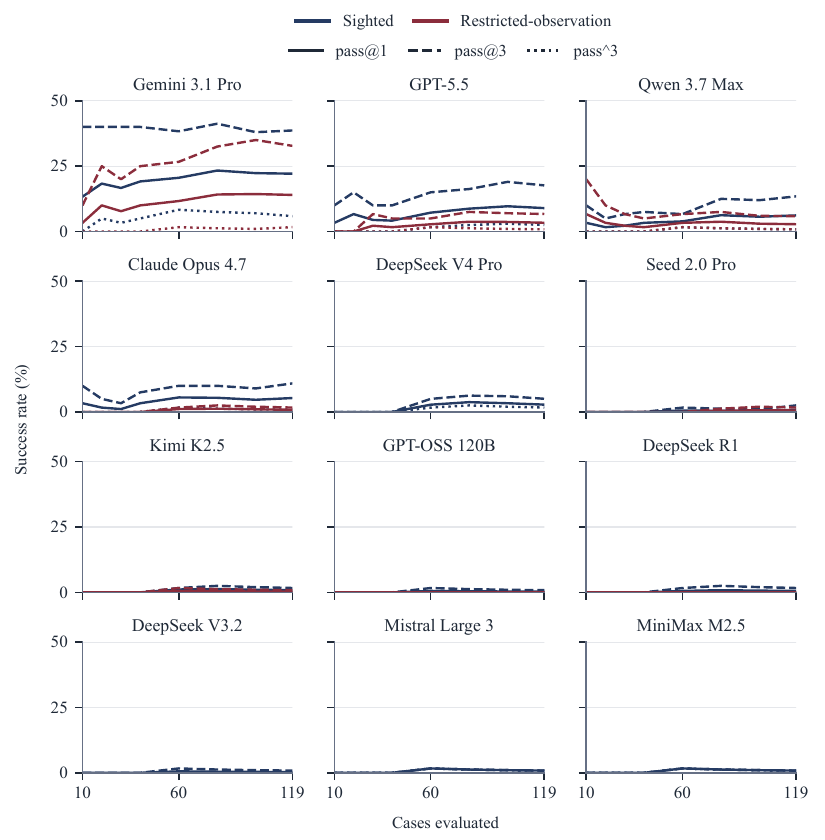}
\caption{\textbf{Estimates on nested case subsets.} Pass@1 (solid), pass@3 (dashed), and \passhat{3}
(dotted) for $m = 10, 20, 30, 40, 60, 80, 100, 119$ cases under one shared random ordering; the full set
reproduces the main results.}
\label{fig:sample-size-stability}
\end{figure}

We recompute all metrics on nested subsets of $m$ cases, keeping three trials per case
(Figure~\ref{fig:sample-size-stability}). Across 1,000 random case orderings, the 2.5th--97.5th percentile
range of Gemini 3.1 Pro's Sighted pass@1 narrows from 3.3--43.3\% at $m = 10$ to 19.7--24.3\% at $m = 100$,
against 22.1\% on all 119 cases. \passhat{3} is the most fragile metric: at $m = 10$ it is zero in 50.6\% of
Gemini's and 75.5\% of GPT-5.5's Sighted subsets. These ranges describe variation within the corpus, not
confidence intervals for new positions.

\FloatBarrier
\section{Observation Ablation}
\label{app:obs-ablation}

Restricted observation changes more than what the agent sees: it removes state refreshes and the query
commands, but it also uses a different system prompt (Appendix~\ref{app:prompt}). To separate these
factors, we ran a $2 \times 2$ ablation after the main evaluation, with a correct check field and a single
endgame prompt whose rules and move-format sections match the Sighted prompt; only the paragraphs
describing observation and commands vary. The factors are \emph{state refreshes} (board, FEN, and legal
moves attached to every message) and \emph{query commands} (\texttt{view\_board}, \texttt{legal\_moves},
\texttt{simulate}). The \textit{Full} arm has both, \textit{No queries} keeps only the refreshes,
\textit{No refresh} keeps only the commands, and \textit{Neither} removes both. Without refreshes, the board
and FEN are shown once and later messages report only the move, capture, check status, and material.
Defender, budgets, and scoring match the main evaluation.

We evaluate Gemini 3.1 Pro and GPT-5.5 on positions each won at least once under Sighted observation, since
a never-solved position scores zero in every arm: 20 of Gemini's 46, drawn by mate distance with a fixed
seed, and all 21 of GPT-5.5's. Each model--position--arm cell has three scored trials, with trials lost to
API errors rerun. GPT-5.5 was served through a different API gateway than in the main evaluation, with the
same model version. Because positions were selected on
main-evaluation wins, we compare arms only within the ablation.

\begin{table}[!htbp]
\centering
\small
\setlength{\tabcolsep}{4pt}
\caption{\textbf{Observation ablation.} A $2\times2$ design crossing per-step state refreshes
with the query commands (\texttt{view\_board}, \texttt{legal\_moves}, \texttt{simulate}); all arms
share one system prompt and a correct check field. Pass@1 is over three trials per position;
Solved counts positions won at least once; Rej.\ is the share of move attempts rejected; the last
three columns give query calls per agent turn. For reference, the main evaluation on the same
positions gives 51.7\% (Sighted) and 20.0\% (Restricted observation) for Gemini 3.1 Pro and
50.8\% and 9.5\% for GPT-5.5; the Sighted rates are biased upward because positions were selected
on Sighted wins.}
\label{tab:obs-ablation}
\begin{tabular}{@{}lccrrrrrr@{}}
\toprule
 & \multicolumn{2}{c}{Factors} & & & & \multicolumn{3}{c}{Queries per turn} \\
\cmidrule(lr){2-3}\cmidrule(l){7-9}
Arm & Refresh & Queries & Pass@1 (\%) & Solved & Rej.\ (\%) & View & Legal & Sim. \\
\midrule
\multicolumn{9}{@{}l}{\textit{Gemini 3.1 Pro, 20 positions}} \\
\quad Full       & \cmark & \cmark & 48.3 & 17/20 & 1.7 & 0.00 & 0.00 & 0.75 \\
\quad No queries & \cmark & \xmark & 33.3 & 11/20 & 0.7 & --   & --   & --   \\
\quad No refresh & \xmark & \cmark & 35.0 & 14/20 & 4.0 & 0.32 & 0.19 & 1.00 \\
\quad Neither    & \xmark & \xmark & 41.7 & 15/20 & 6.9 & --   & --   & --   \\
\midrule
\multicolumn{9}{@{}l}{\textit{GPT-5.5, 21 positions}} \\
\quad Full       & \cmark & \cmark & 39.7 & 15/21 & 0.3 & 0.01 & 0.05 & 0.98 \\
\quad No queries & \cmark & \xmark & 25.4 & 10/21 & 1.9 & --   & --   & --   \\
\quad No refresh & \xmark & \cmark & 34.9 & 13/21 & 2.0 & 0.38 & 0.67 & 1.03 \\
\quad Neither    & \xmark & \xmark & 38.1 & 15/21 & 3.6 & --   & --   & --   \\
\bottomrule
\end{tabular}
\end{table}

Intervals below are 95\% case-bootstrap intervals and $p$-values are exact sign tests over positions.
(i)~Removing the query commands costs both models about 15 points (\textit{Full} minus \textit{No queries}:
15.0 for Gemini, [0.0, 30.0]; 14.3 for GPT-5.5, [1.6, 23.8], $p = 0.013$); the agents lose simulation, which
they use on most turns. (ii)~Removing the refreshes while keeping the commands costs Gemini 13.3 points
([0.0, 25.0], $p = 0.039$) and GPT-5.5 4.8 ([$-$11.1, 19.0]); both then query the state, which they almost
never do when it is refreshed. (iii)~With neither, the two models win 41.7\% and 38.1\% of trials, close to
\textit{Full} (differences of 6.7 and 1.6 points, intervals spanning zero) and far above the 20.0\% and 9.5\%
they reach on the same positions under Restricted observation. Without refreshes, removing the commands
does not lower pass@1 further, so the command effect in (i) appears only when the state is refreshed; with
three trials per position we do not interpret this interaction further. Most of the Restricted drop in
Section~\ref{sec:results} therefore reflects the setting as a whole, including its prompt and feedback,
rather than the withheld board alone.

\FloatBarrier
\section{Behavioral Diagnostics}
\label{app:behavior}

The analyses in this section are descriptive. Models, positions, and decision states vary together, so
the contrasts below characterize behavior without identifying its causes.

\subsection{Setting and Failure Profiles}
\label{app:setting-diagnostics}

\begin{figure}[!htbp]
\centering
\includegraphics[width=\linewidth]{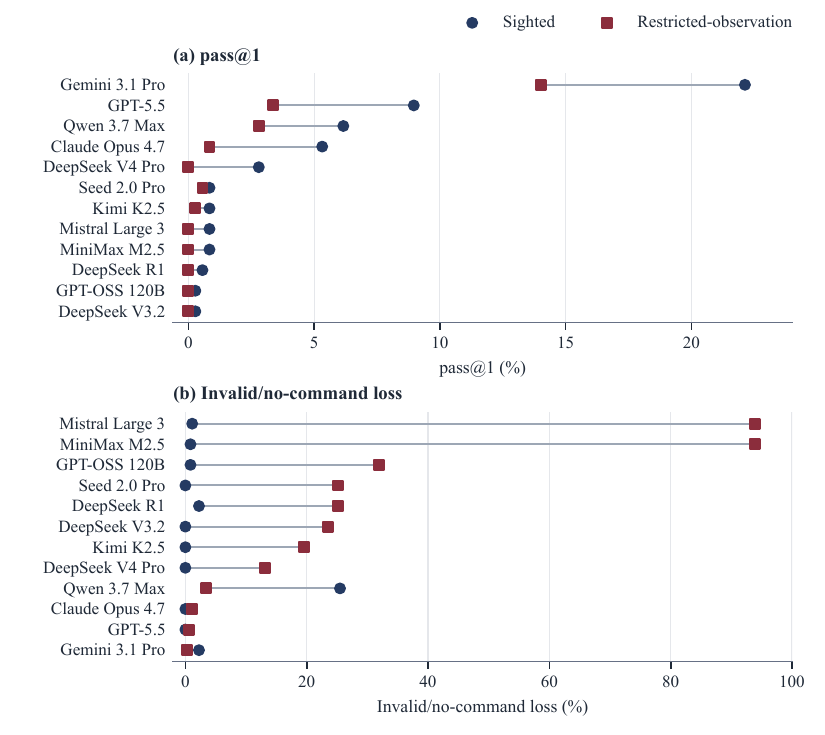}
\caption{\textbf{The setting difference varies across models.} (a) Pass@1; (b) rate of terminal
invalid-action or no-command losses, each over 357 trials per model and setting.}
\label{fig:protocol-appendix}
\end{figure}

\begin{figure}[!htbp]
\centering
\includegraphics[width=\linewidth]{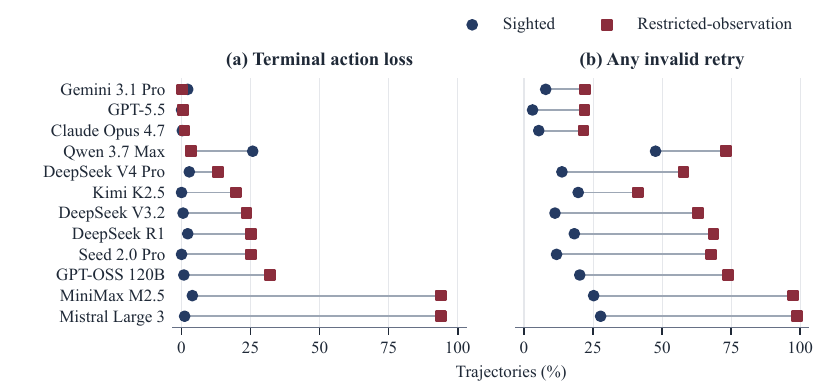}
\caption{\textbf{Action-level failures concentrate under Restricted observation.} Left: terminal action
losses (invalid action, no command, or action limit). Right: trajectories with any rejected move, which
may be followed by recovery.}
\label{fig:failure-patterns-appendix}
\end{figure}

Figure~\ref{fig:game-length} in Section~\ref{sec:simulation-analysis} shows game length. No win ends before
the stored mate distance, and the median failed trial lasts six plies under Sighted and four under
Restricted observation. Most failures are lost on the board (76.1\% Sighted, 66.0\% Restricted).
Resignation ends 18.4\% of failed Sighted trials (40.3\% for Gemini 3.1 Pro) but only 3.2\% of Restricted
ones, where action forfeits end 28.2\%; Mistral Large 3 and MiniMax M2.5 forfeit 310 and 313 of their 357
Restricted trials before playing a move.

\FloatBarrier
\subsection{Simulation and First-Move Choices}
\label{app:tool-diagnostics}

Tables~\ref{tab:simulation-overview}--\ref{tab:turn-patterns} break down simulation by setting, model,
request length, and turn pattern, and Table~\ref{tab:failure-patterns} separates terminal action losses
from recoverable rejections. GPT-5.5 mostly submits one-ply simulations, whereas several other models
submit multi-ply lines. Complete-line legality falls with length, from 96.3\% (2,496/2,592) for one ply to
47.6\% (1,110/2,332) for three or four plies and 37.4\% (357/955) for five to eight, although length,
model, and position vary together. Tables~\ref{tab:decision-repl-sighted}--\ref{tab:piece-choice-repl-blind}
profile the first real move against the uniform baseline over legal moves; for models that often forfeit
early under Restricted observation, these profiles condition on trajectories with an accepted first move.

\begin{figure}[!htbp]
\centering
\includegraphics[width=\linewidth]{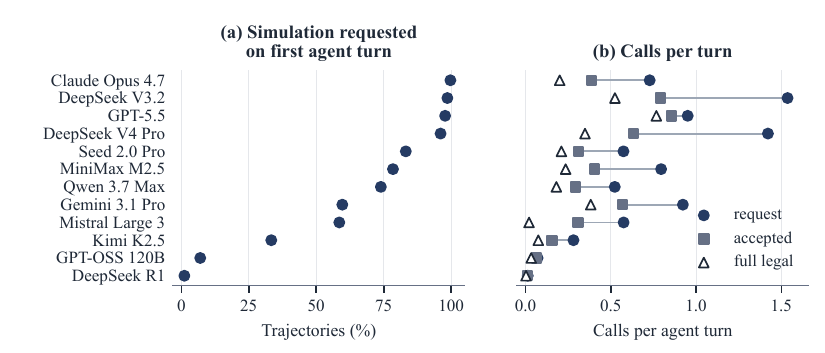}
\caption{\textbf{Simulation use varies widely across models.} (a) Trajectories that request simulation on
the first turn (357 per model). (b) Simulation requests, accepted calls, and complete legal lines per agent
turn over 15,318 agent turns; requests include budget-rejected calls.}
\label{fig:tool-patterns}
\end{figure}

% Generated by scripts/analyze_case_behaviors.py; do not edit manually.
\begin{table}[t]
\centering
\caption{Explicit simulation use by protocol. All entries count trajectories. Behavior rows overlap; acceptance of a simulation call does not imply full-line legality.}
\label{tab:simulation-overview}
\fontsize{9}{11}\selectfont
\setlength{\tabcolsep}{4pt}
\begin{tabular}{lrr}
\toprule
Trajectory measure & Sighted & Restricted-observation \\
\midrule
Trajectory denominator & 4284 & 4284 \\
Requested simulation & 3098 & 0 \\
At least one accepted simulation call & 3098 & 0 \\
At least one full legal requested line & 2155 & 0 \\
At least one mid-line invalid step & 1639 & 0 \\
At least one capped request & 32 & 0 \\
Only budget/protocol rejected & 0 & 0 \\
\bottomrule
\end{tabular}
\end{table}

% Generated by scripts/analyze_case_behaviors.py; do not edit manually.
\begin{table}[t]
\centering
\caption{Sighted explicit simulation use by model. Any and Full count trajectories out of 357: any request and at least one full legal original request, respectively. Accepted and Rejected count calls; all rejections here are budget rejections.}
\label{tab:simulation-by-alias}
\fontsize{9}{11}\selectfont
\setlength{\tabcolsep}{4pt}
\begin{tabular}{lrrrr}
\toprule
Model & Any & Full & Accepted & Rejected \\
\midrule
Gemini 3.1 Pro & 332 & 299 & 1081 & 675 \\
GPT-5.5 & 356 & 353 & 1727 & 191 \\
Qwen 3.7 Max & 290 & 188 & 360 & 284 \\
Claude Opus 4.7 & 357 & 241 & 595 & 529 \\
DeepSeek V4 Pro & 354 & 266 & 810 & 1006 \\
Seed 2.0 Pro & 328 & 222 & 426 & 359 \\
Kimi K2.5 & 156 & 76 & 180 & 148 \\
GPT-OSS 120B & 63 & 33 & 70 & 7 \\
DeepSeek R1 & 8 & 2 & 9 & 2 \\
DeepSeek V3.2 & 352 & 292 & 890 & 835 \\
Mistral Large 3 & 213 & 14 & 242 & 212 \\
MiniMax M2.5 & 289 & 169 & 359 & 345 \\
\bottomrule
\end{tabular}
\end{table}

% Generated by scripts/analyze_case_behaviors.py; do not edit manually.
\begin{table}[t]
\centering
\caption{Original request length among accepted Sighted simulation calls. Length is measured in plies. Two or more plies include model-supplied assumed opponent replies, not engine search. Length and position differences prevent interpreting legality rates as planning ability.}
\label{tab:simulation-length}
\fontsize{9}{11}\selectfont
\setlength{\tabcolsep}{4pt}
\begin{tabular}{lrrrr}
\toprule
Model & Median & Mean & 1 ply (\%) & $\geq$2 plies (\%) \\
\midrule
Gemini 3.1 Pro & 3 & 3.57 & 12.5 & 87.5 \\
GPT-5.5 & 1 & 1.52 & 71.9 & 28.1 \\
Qwen 3.7 Max & 4 & 4.20 & 9.2 & 90.8 \\
Claude Opus 4.7 & 3 & 3.38 & 5.5 & 94.5 \\
DeepSeek V4 Pro & 3 & 3.12 & 27.0 & 73.0 \\
Seed 2.0 Pro & 2 & 2.20 & 46.7 & 53.3 \\
Kimi K2.5 & 3 & 3.33 & 24.4 & 75.6 \\
GPT-OSS 120B & 1.5 & 1.86 & 50.0 & 50.0 \\
DeepSeek R1 & 3 & 3.44 & 22.2 & 77.8 \\
DeepSeek V3.2 & 2 & 1.92 & 48.4 & 51.6 \\
Mistral Large 3 & 3 & 3.93 & 3.7 & 96.3 \\
MiniMax M2.5 & 1 & 1.68 & 58.5 & 41.5 \\
\bottomrule
\end{tabular}
\end{table}

% Generated by scripts/analyze_case_behaviors.py; do not edit manually.
\begin{table}[t]
\centering
\caption{Mutually exclusive agent-turn patterns, assigned by the first applicable row in the displayed priority order. Entries are counts (percent), with 15,318 Sighted and 12,387 Restricted-observation agent turns. Recovery means only that a real move followed a rejected or aborted simulation, not that the strategy improved. Direct execution excludes explicit auxiliary actions but does not imply the absence of automatically supplied state information.}
\label{tab:turn-patterns}
\fontsize{9}{11}\selectfont
\setlength{\tabcolsep}{4pt}
\begin{tabular}{@{}p{0.49\columnwidth}rr@{}}
\toprule
Turn pattern (priority order) & Sighted $n$ (\%) & Restricted-observation $n$ (\%) \\
\midrule
No executed move & 898 (5.9) & 1,319 (10.6) \\
Simulation rejected/aborted, then recovery & 4,386 (28.6) & 0 (0.0) \\
Simulate, then adopt tested root & 1,447 (9.4) & 0 (0.0) \\
Simulate, then choose different root & 774 (5.1) & 0 (0.0) \\
Observe, then move & 234 (1.5) & 15 (0.1) \\
Direct execution & 6,988 (45.6) & 10,581 (85.4) \\
Other explicit action, then move & 591 (3.9) & 472 (3.8) \\
\bottomrule
\end{tabular}
\end{table}

% Generated by scripts/analyze_case_behaviors.py; do not edit manually.
\begin{table}[t]
\centering
\caption{Mutually exclusive observed failure classes. Opponent terminal collects the remaining failures; on replay, each ends with the attacker checkmated or stalemated. Classes follow the released long-check rule (Appendix~\ref{app:rules-check}).}
\label{tab:failure-patterns}
\fontsize{9}{11}\selectfont
\setlength{\tabcolsep}{4pt}
\begin{tabular}{lrr}
\toprule
Observed failure class & Sighted & Restricted-observation \\
\midrule
All observed failures & 4106 & 4206 \\
Invalid-action limit & 12 & 1178 \\
No command/tool call & 105 & 6 \\
Perpetual check by the attacker & 16 & 7 \\
Repetition draw & 39 & 56 \\
Action limit & 25 & 0 \\
Turn cap & 29 & 47 \\
Resignation & 756 & 135 \\
Opponent terminal (attacker cannot move) & 3124 & 2777 \\
\bottomrule
\end{tabular}
\end{table}

% Generated by scripts/analyze_decision_styles.py.
\begin{table}[t]
\centering
\caption{Root-action composition in the Sighted setting. Entries are observed percent (percentage-point difference from uniform choice among true legal root moves). For each model, the reference uses exactly the cases with an accepted first move. Check and capture overlap; quiet means neither. These are action patterns, not personality labels.}
\label{tab:decision-repl-sighted}
\fontsize{9}{11}\selectfont
\setlength{\tabcolsep}{4pt}
\begin{tabular}{lrrrr}
\toprule
Model & Accepted & Check & Capture & Quiet \\
\midrule
Gemini 3.1 Pro & 353 & 95.5 (+84.8) & 25.5 (+18.7) & 4.2 (-80.8) \\
GPT-5.5 & 357 & 95.8 (+85.1) & 25.5 (+18.7) & 3.6 (-81.3) \\
Qwen 3.7 Max & 287 & 88.9 (+77.9) & 27.2 (+20.3) & 8.7 (-76.0) \\
Claude Opus 4.7 & 356 & 84.6 (+73.8) & 29.8 (+23.0) & 10.7 (-74.3) \\
DeepSeek V4 Pro & 350 & 81.4 (+70.7) & 27.7 (+21.0) & 14.6 (-70.4) \\
Seed 2.0 Pro & 357 & 88.0 (+77.2) & 33.3 (+26.6) & 7.3 (-77.7) \\
Kimi K2.5 & 357 & 72.8 (+62.1) & 26.3 (+19.5) & 19.9 (-65.1) \\
GPT-OSS 120B & 352 & 71.9 (+61.2) & 20.7 (+14.0) & 22.2 (-62.8) \\
DeepSeek R1 & 352 & 51.4 (+40.7) & 34.7 (+27.8) & 34.7 (-50.3) \\
DeepSeek V3.2 & 355 & 70.4 (+59.7) & 41.1 (+34.3) & 15.8 (-69.2) \\
Mistral Large 3 & 355 & 37.5 (+26.7) & 25.6 (+18.8) & 46.5 (-38.5) \\
MiniMax M2.5 & 344 & 34.9 (+24.2) & 27.0 (+20.3) & 52.6 (-32.4) \\
\bottomrule
\end{tabular}
\end{table}

\begin{table}[t]
\centering
\caption{Root piece-choice differences (percentage points) in the Sighted setting. Differences are relative to uniform legal choice on the same accepted-first-move case distribution. Other comprises general, advisor, and elephant. Intervals are omitted for readability, and no multiple-comparison correction is applied. These exploratory contrasts are not evidence of piece-specific playing strength.}
\label{tab:piece-choice-repl-sighted}
\fontsize{9}{11}\selectfont
\setlength{\tabcolsep}{4pt}
\begin{tabular}{lrrrrr}
\toprule
Model & Rook & Cannon & Horse & Pawn & Other \\
\midrule
Gemini 3.1 Pro & +10.9 & -22.5 & +3.2 & +10.8 & -2.5 \\
GPT-5.5 & +16.1 & -17.5 & -5.7 & +8.7 & -1.6 \\
Qwen 3.7 Max & +4.7 & -17.8 & +6.2 & +8.8 & -1.9 \\
Claude Opus 4.7 & +4.7 & -22.2 & +11.2 & +9.3 & -3.0 \\
DeepSeek V4 Pro & +11.7 & -21.9 & +7.7 & +4.5 & -1.9 \\
Seed 2.0 Pro & +24.2 & -27.3 & -4.0 & +9.9 & -2.7 \\
Kimi K2.5 & +4.3 & -16.9 & +6.9 & +8.7 & -3.0 \\
GPT-OSS 120B & +16.5 & -16.6 & +5.4 & -2.7 & -2.6 \\
DeepSeek R1 & +17.8 & -22.3 & +2.1 & +3.5 & -1.1 \\
DeepSeek V3.2 & +11.8 & -21.6 & +2.5 & +9.9 & -2.7 \\
Mistral Large 3 & +1.7 & -12.0 & +14.7 & -1.6 & -2.8 \\
MiniMax M2.5 & +12.0 & -25.8 & +11.9 & +3.2 & -1.4 \\
\bottomrule
\end{tabular}
\end{table}

\begin{table}[t]
\centering
\caption{Root-action composition in the Restricted-observation setting. Entries are observed percent (percentage-point difference from uniform choice among true legal root moves). For each model, the reference uses exactly the cases with an accepted first move. Check and capture overlap; quiet means neither. These are action patterns, not personality labels.}
\label{tab:decision-repl-blind}
\fontsize{9}{11}\selectfont
\setlength{\tabcolsep}{4pt}
\begin{tabular}{lrrrr}
\toprule
Model & Accepted & Check & Capture & Quiet \\
\midrule
Gemini 3.1 Pro & 356 & 82.6 (+71.9) & 26.7 (+19.9) & 10.4 (-74.6) \\
GPT-5.5 & 357 & 72.5 (+61.8) & 37.8 (+31.0) & 13.7 (-71.3) \\
Qwen 3.7 Max & 352 & 69.9 (+59.1) & 26.7 (+19.9) & 18.8 (-66.2) \\
Claude Opus 4.7 & 355 & 58.9 (+48.1) & 45.1 (+38.3) & 19.7 (-65.2) \\
DeepSeek V4 Pro & 341 & 43.4 (+32.7) & 36.1 (+29.3) & 34.0 (-50.9) \\
Seed 2.0 Pro & 334 & 55.7 (+45.0) & 35.3 (+28.7) & 32.0 (-53.0) \\
Kimi K2.5 & 340 & 52.1 (+41.4) & 46.5 (+39.6) & 25.3 (-59.7) \\
GPT-OSS 120B & 320 & 61.3 (+50.7) & 41.9 (+35.0) & 24.1 (-61.0) \\
DeepSeek R1 & 325 & 51.7 (+40.8) & 42.5 (+35.7) & 28.3 (-56.5) \\
DeepSeek V3.2 & 303 & 36.0 (+25.3) & 29.4 (+22.9) & 50.2 (-35.1) \\
Mistral Large 3 & 47 & 12.8 (+5.1) & 0.0 (-3.7) & 87.2 (-2.7) \\
MiniMax M2.5 & 44 & 27.3 (+18.9) & 4.5 (-0.5) & 72.7 (-15.7) \\
\bottomrule
\end{tabular}
\end{table}

\begin{table}[t]
\centering
\caption{Root piece-choice differences (percentage points) in the Restricted-observation setting. Differences are relative to uniform legal choice on the same accepted-first-move case distribution. Other comprises general, advisor, and elephant. Intervals are omitted for readability, and no multiple-comparison correction is applied. These exploratory contrasts are not evidence of piece-specific playing strength.}
\label{tab:piece-choice-repl-blind}
\fontsize{9}{11}\selectfont
\setlength{\tabcolsep}{4pt}
\begin{tabular}{lrrrrr}
\toprule
Model & Rook & Cannon & Horse & Pawn & Other \\
\midrule
Gemini 3.1 Pro & +11.4 & -19.4 & +2.5 & +7.6 & -2.2 \\
GPT-5.5 & +11.6 & -17.5 & -1.2 & +8.2 & -1.1 \\
Qwen 3.7 Max & +0.5 & -15.5 & +5.2 & +9.2 & +0.6 \\
Claude Opus 4.7 & +5.4 & -27.8 & +10.4 & +13.0 & -1.1 \\
DeepSeek V4 Pro & +7.7 & -19.5 & +4.4 & +6.3 & +1.1 \\
Seed 2.0 Pro & +18.7 & -16.1 & -6.8 & +7.0 & -2.8 \\
Kimi K2.5 & +4.9 & -21.5 & +4.3 & +13.8 & -1.6 \\
GPT-OSS 120B & +18.9 & -24.1 & -0.5 & +7.6 & -1.8 \\
DeepSeek R1 & +25.2 & -25.6 & -1.3 & +2.8 & -1.1 \\
DeepSeek V3.2 & +7.0 & -22.6 & +8.4 & +2.4 & +4.8 \\
Mistral Large 3 & -2.3 & +5.9 & -5.5 & +3.9 & -1.9 \\
MiniMax M2.5 & +1.2 & -4.4 & -1.0 & +2.2 & +2.0 \\
\bottomrule
\end{tabular}
\end{table}

\FloatBarrier
\subsection{REPL-Call Composition by Setting, Outcome, and Model}
\label{app:action-phases}

\begin{figure}[!htbp]
\centering
\includegraphics[width=\linewidth]{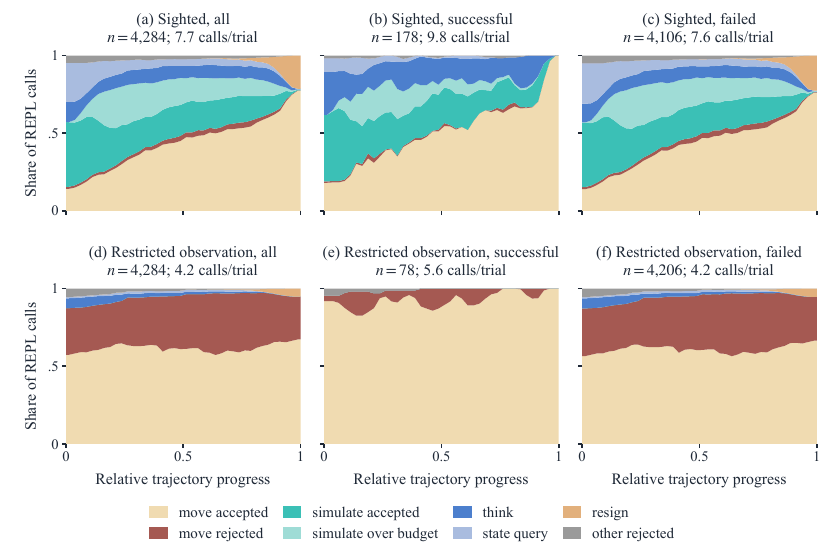}
\caption{\textbf{REPL-call composition by setting and outcome}, pooling all 12 models. Columns show all,
successful, and failed trials.}
\label{fig:action-phases-overview}
\end{figure}

\begin{figure}[!htbp]
\centering
\includegraphics[width=\linewidth]{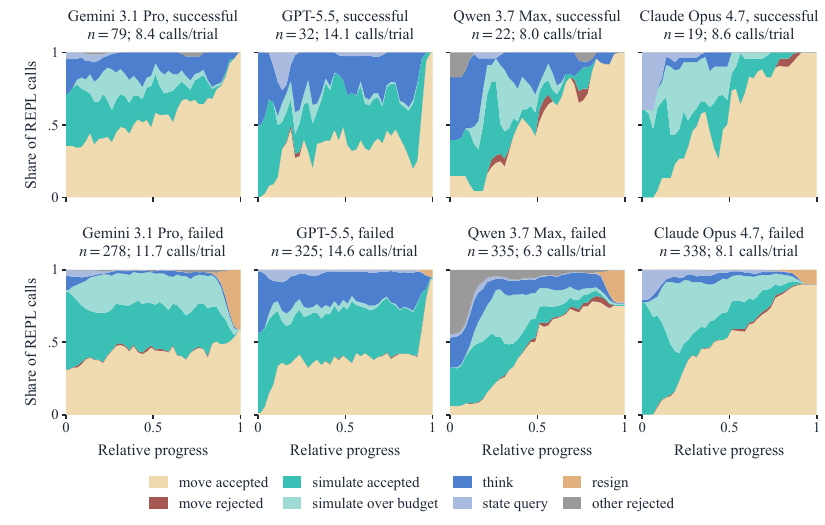}
\caption{\textbf{Sighted REPL-call composition by outcome for the four models with the most wins.}
Successful panels contain 19--79 trials.}
\label{fig:action-phases-outcome-models}
\end{figure}

\begin{figure}[!htbp]
\centering
\includegraphics[width=\linewidth]{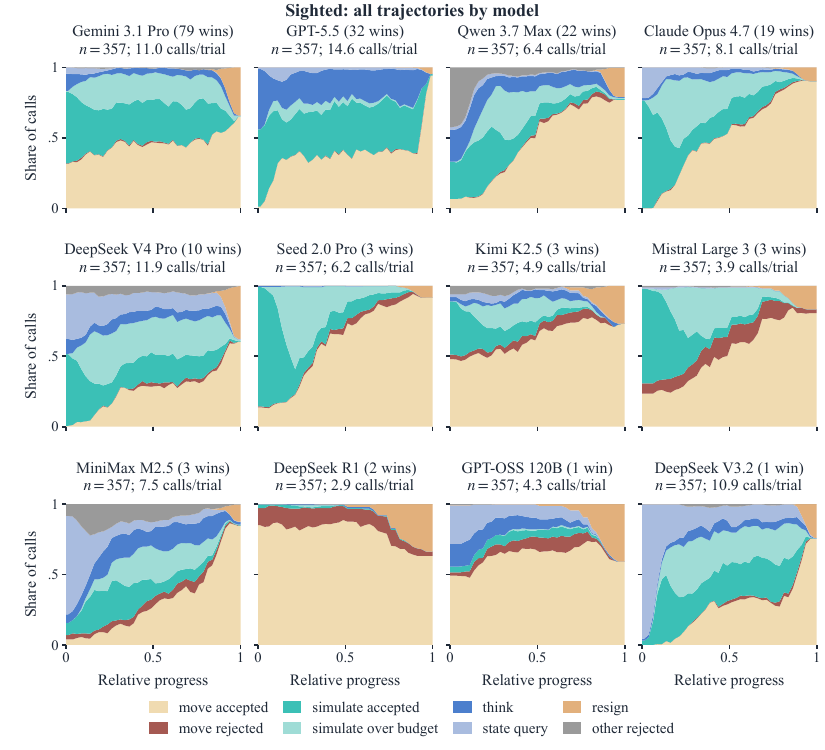}
\caption{\textbf{Sighted REPL-call composition for each model} over all 357 trials.}
\label{fig:action-phases-models-sighted}
\end{figure}

\begin{figure}[!htbp]
\centering
\includegraphics[width=\linewidth]{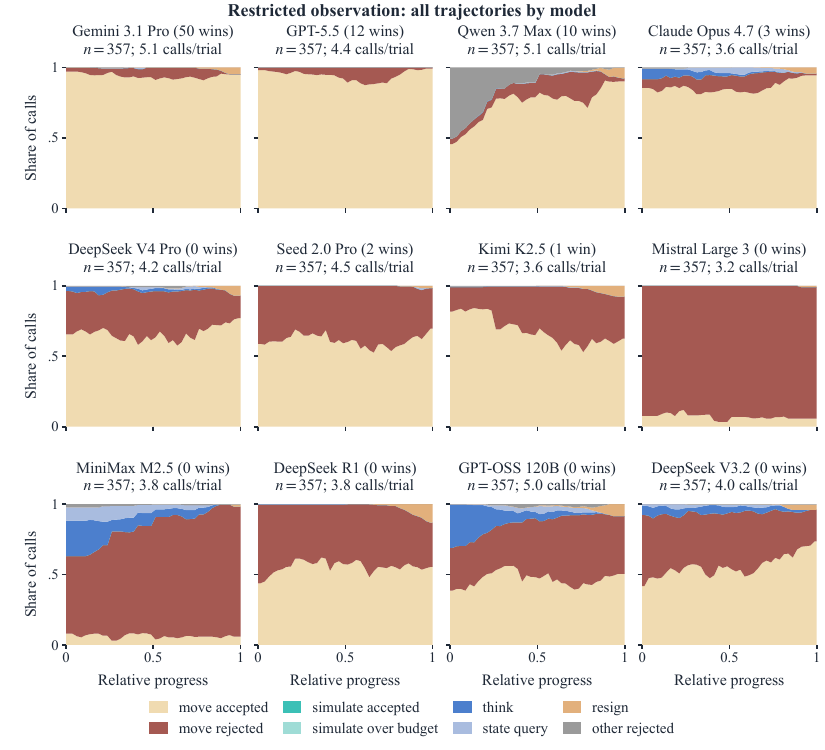}
\caption{\textbf{Restricted-observation REPL-call composition for each model} over all 357 trials. State
queries here are \texttt{get\_history} calls, the only query available.}
\label{fig:action-phases-models-restricted}
\end{figure}

Figures~\ref{fig:action-phases-overview}--\ref{fig:action-phases-models-restricted} apply the construction
of Figure~\ref{fig:action-phases} by setting and outcome, to the four Sighted models with the most wins,
and to every model in each setting, excluding environment-generated forfeit records.

\FloatBarrier
\subsection{Feedback and Error Recovery}
\label{app:behavior-events}

\begin{figure}[!htbp]
\centering
\includegraphics[width=\linewidth]{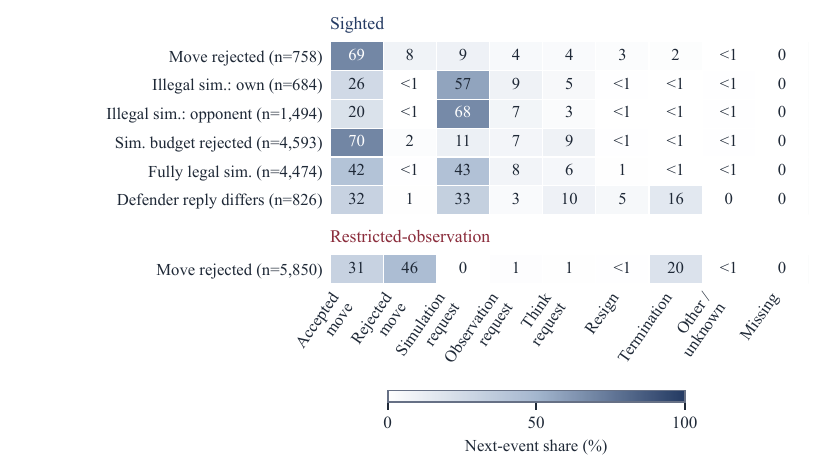}
\caption{\textbf{Next action after selected feedback events.} Each row conditions on one event and shows
the next interaction in the same turn; for defender-reply mismatch, the next decision after the real reply.
Requests include rejected ones, and percentages are rounded.}
\label{fig:feedback-transitions}
\end{figure}

\begin{figure}[!htbp]
\centering
\includegraphics[width=\linewidth]{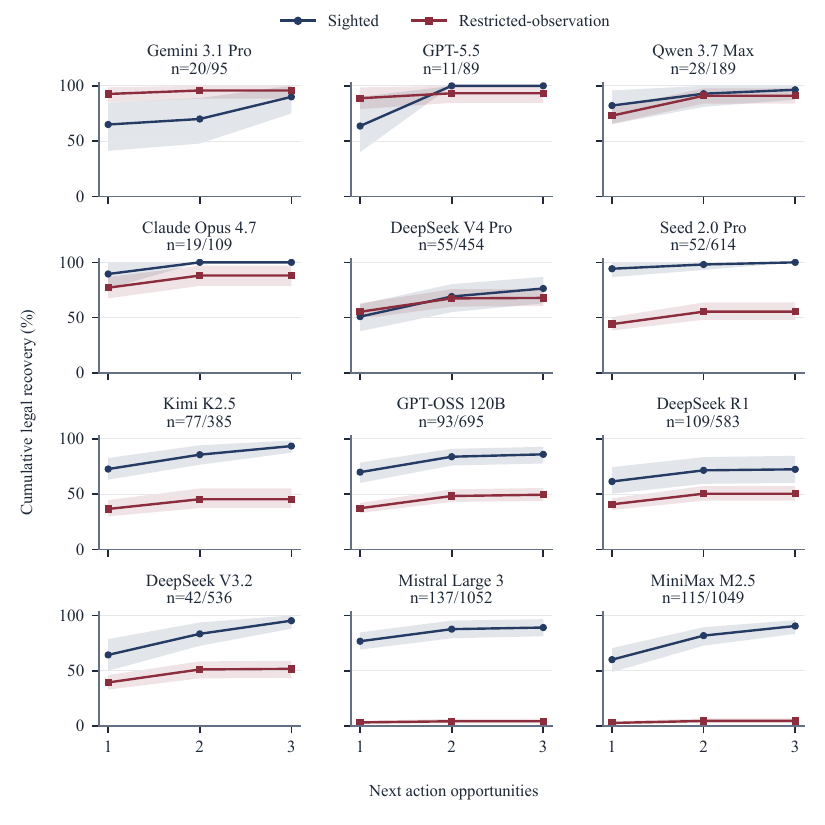}
\caption{\textbf{Recovery after a rejected move, by model.} Cumulative fraction of rejected moves followed
by a legal real move within one, two, or three action interactions of the same turn; $n$ gives
Sighted/Restricted rejections. Shading: 95\% case-bootstrap intervals (1,000 resamples).}
\label{fig:error-recovery}
\end{figure}

Figure~\ref{fig:feedback-transitions} shows how agents respond to feedback. For recovery
(Figure~\ref{fig:error-recovery}), a rejected real move counts as recovered if rules replay confirms a legal
real move within the next one, two, or three action interactions of the same turn; reads, simulations, and
\texttt{think} calls occupy window positions, and resignation and forfeit compete with recovery. Under
Sighted observation, 69.4\%, 83.1\%, and 87.9\% of the 758 rejections are recovered within one, two, and
three interactions; under Restricted observation, 31.2\%, 38.6\%, and 38.9\% of the 5,850 rejections are
(three-interaction intervals [84.6, 90.9]\% and [36.0, 42.1]\%). The Restricted total is dominated by
Mistral Large 3 and MiniMax M2.5, which contribute 1,052 and 1,049 rejections and recover from 4.3\% and
4.6\% of them. The four strongest models recover from at least 88\% in both settings (Gemini 3.1 Pro 90.0\%
Sighted and 95.8\% Restricted; GPT-5.5 100\% and 93.3\%; Claude Opus 4.7 100\% and 88.1\%; Qwen 3.7 Max
96.4\% and 91.0\%), though each has only 11--28 Sighted rejections.

\FloatBarrier
\subsection{Piece Composition}

\begin{figure}[htbp]
\centering
\includegraphics[width=\linewidth]{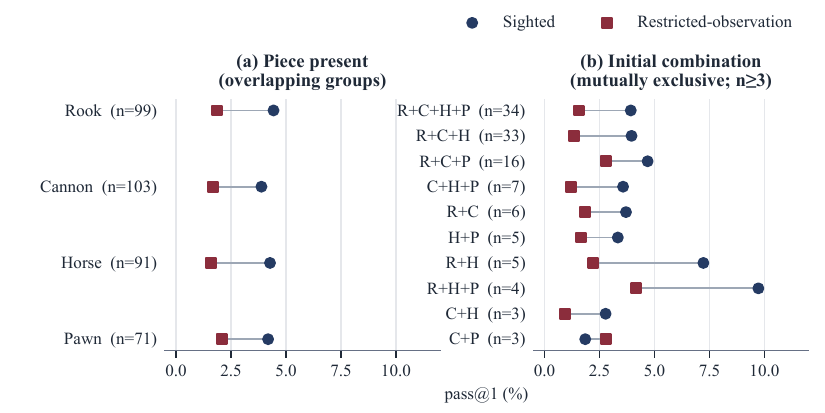}
\caption{\textbf{Outcomes by initial attacking pieces.} Left: overlapping rook, cannon, horse, and pawn
presence groups. Right: mutually exclusive combinations with at least three cases. Labels give case
counts; points pool 12 models and three trials per case.}
\label{fig:piece-groups-appendix}
\end{figure}

Figure~\ref{fig:piece-groups-appendix} and
Tables~\ref{tab:piece-presence}--\ref{tab:first-move-piece} group cases by initial attacking material.
Presence groups overlap, combination groups are mutually exclusive, and first-move counts reflect piece
availability and trajectory length rather than piece preference.

% Generated by scripts/analyze_case_behaviors.py; do not edit manually.
\begin{table}[t]
\centering
\caption{Observed passes by initial red-piece presence. Rows overlap and must not be added. Position structure and difficulty confound comparisons; these are not causal piece effects.}
\label{tab:piece-presence}
\fontsize{9}{11}\selectfont
\setlength{\tabcolsep}{4pt}
\begin{tabular}{lrrr}
\toprule
Initial red piece & Cases & Sighted pass/traj. & Restricted-observation pass/traj. \\
\midrule
Rook & 99 & 158/3564 & 66/3564 \\
Cannon & 103 & 144/3708 & 62/3708 \\
Horse & 91 & 140/3276 & 52/3276 \\
Pawn & 71 & 107/2556 & 53/2556 \\
\bottomrule
\end{tabular}
\end{table}

% Generated by scripts/analyze_case_behaviors.py; do not edit manually.
\begin{table}[t]
\centering
\caption{Mutually exclusive initial red-piece presence combinations. Groups with fewer than five cases are retained for coverage, not model rankings.}
\label{tab:piece-combinations}
\fontsize{9}{11}\selectfont
\setlength{\tabcolsep}{4pt}
\begin{tabular}{lr}
\toprule
Combination & Cases \\
\midrule
rook+cannon+horse+pawn & 34 \\
rook+cannon+horse & 33 \\
rook+cannon+pawn & 16 \\
cannon+horse+pawn & 7 \\
rook+cannon & 6 \\
horse+pawn & 5 \\
rook+horse & 5 \\
rook+horse+pawn & 4 \\
cannon+horse & 3 \\
cannon+pawn & 3 \\
cannon & 1 \\
pawn & 1 \\
rook+pawn & 1 \\
\bottomrule
\end{tabular}
\end{table}

% Generated by scripts/analyze_case_behaviors.py; do not edit manually.
\begin{table}[t]
\centering
\caption{Piece type of the first accepted non-synthetic agent move, over 4,284 trajectories per setting. Other comprises general, advisor, and elephant; unknown denotes unparsed feedback. This is action composition, not position-independent preference.}
\label{tab:first-move-piece}
\fontsize{9}{11}\selectfont
\setlength{\tabcolsep}{4pt}
\begin{tabular}{lrr}
\toprule
First moved piece & Sighted & Restricted-observation \\
\midrule
Rook & 2032 & 1665 \\
Cannon & 705 & 588 \\
Horse & 851 & 608 \\
Pawn & 556 & 525 \\
Other & 31 & 88 \\
Unknown & 0 & 0 \\
No Accepted First Move & 109 & 810 \\
\bottomrule
\end{tabular}
\end{table}

\FloatBarrier
\section{Representative Trajectories}
\label{app:trajectories}

The examples below are selected by fixed, deterministic rules over the retained trajectories. They
illustrate how play unfolds; they are not estimates of model tendencies.

\begin{figure}[!htbp]
\centering
\includegraphics[width=\linewidth]{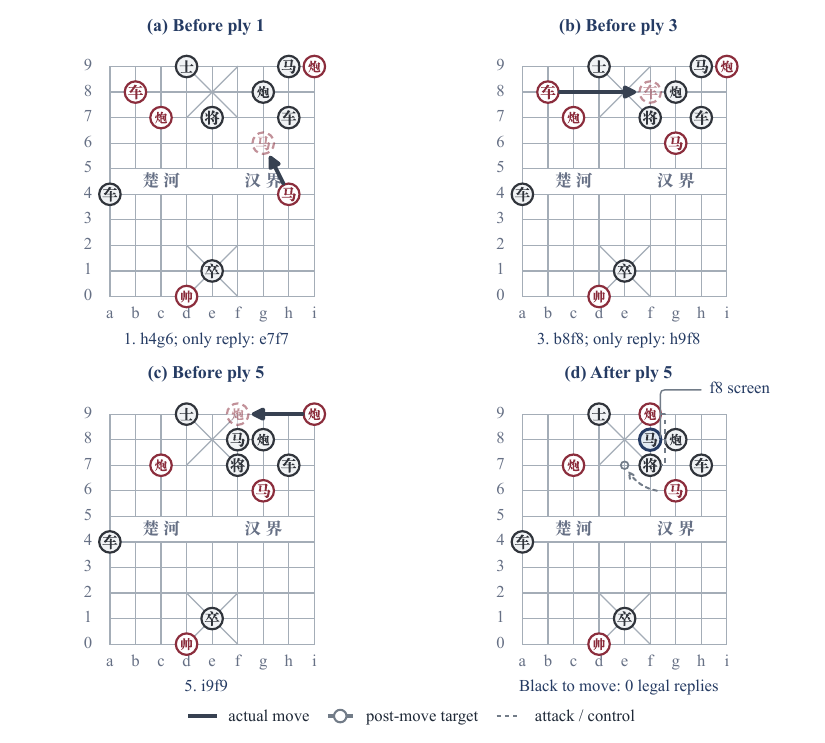}
\caption{\textbf{Forced mate on SQYQ-251.} Panels A--C show the real position before each of Red's three
moves (solid arrow: move played; dashed outline: its destination). Black's replies \texttt{e7f7} and
\texttt{h9f8} are forced. Panel D shows the final position: the cannon on f9 mates through the Black horse
on f8 while the Red horse on g6 covers e7. The archived feedback labelled this terminal a stalemate; rules
replay confirms checkmate.}
\label{fig:tactical-case}
\end{figure}

\paragraph{SQYQ-251: two routes to the same mate.}
\label{app:gemini-overview}
The solution is a forced mate in five plies (Figure~\ref{fig:tactical-case}). After \texttt{h4g6}, the Red
horse controls e7 and protects f8. The rook sacrifice \texttt{b8f8} forces \texttt{h9f8}; the capturing
horse opens the cannon's line from i9 and becomes the screen for \texttt{i9f9}. Qwen 3.7 Max (Sighted,
trial~2) states this plan, simulates the full line \texttt{h4g6 e7f7 b8f8 h9f8 i9f9} at its first decision,
and the real game follows it exactly. Gemini 3.1 Pro (Sighted, trial~3; Figure~\ref{fig:overview})
simulates only after seeing the defender's first reply, checking \texttt{b8f8 h9f8 i9f9} at its second
decision before playing it. On the same position, DeepSeek V4 Pro reaches the prefix \texttt{h4g6 e7f7}
and then plays \texttt{b8e8}, stating that the move gives check. It does not; Black answers \texttt{h7h0}
and mates. The false claim is checkable against the rules, although we do not identify \texttt{b8e8} as
the engine-determined losing move.

\begin{figure}[!htbp]
\centering
\includegraphics[width=\linewidth]{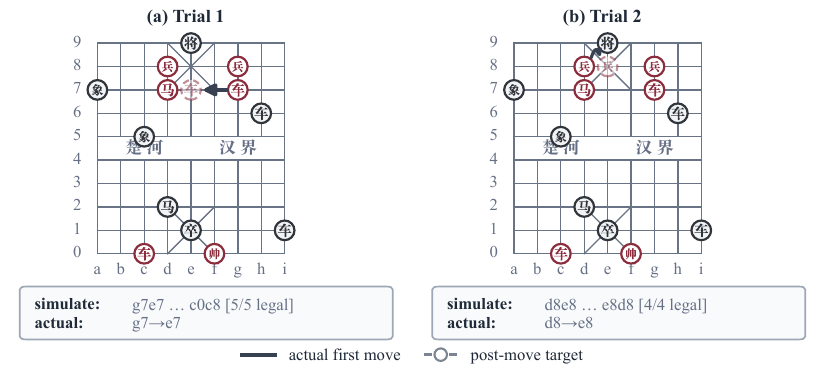}
\caption{\textbf{Same model and position, different first moves.} Two Sighted Gemini 3.1 Pro trials on
SQYQ-012. Trial~1 simulates a legal five-ply line, plays \texttt{g7e7}, and wins; trial~2 simulates a legal
four-ply line, plays \texttt{d8e8}, and resigns. Solid arrows mark first moves; dashed outlines mark their
destinations.}
\label{fig:case-trajectory}
\end{figure}

\paragraph{SQYQ-012: legal lines, different outcomes.} Both Gemini 3.1 Pro trials simulate a fully legal
line before moving (Figure~\ref{fig:case-trajectory}). Trial~1 checks \texttt{g7e7 c5e7 d8e8 e9e8 c0c8},
plays \texttt{g7e7}, and later wins through \texttt{d8e8} and \texttt{d7b8}; trial~2 checks
\texttt{d8e8 e9e8 g8f8 e8d8}, plays \texttt{d8e8}, and eventually resigns. Legality of the simulated line
does not separate the two outcomes.

\begin{figure}[!htbp]
\centering
\includegraphics[width=\linewidth]{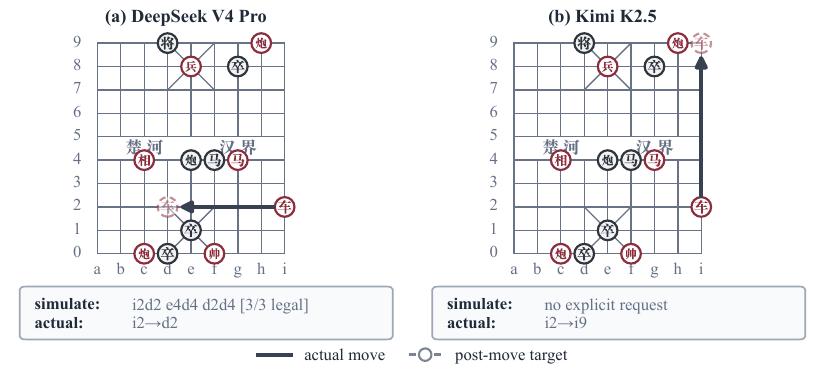}
\caption{\textbf{Different models from the same state on JH-327.} After the shared prefix
\texttt{h2h9 g9g8}, DeepSeek V4 Pro (trial~3) simulates a fully legal three-ply line and plays
\texttt{i2d2} (win); Kimi K2.5 (trial~2) plays \texttt{i2i9} without simulating (loss). Arrows and outlines
follow Figure~\ref{fig:case-trajectory}; simulation details appear in each panel.}
\label{fig:cross-model-case}
\end{figure}

\paragraph{JH-327: same state, different models.} From an identical second decision, the model that
simulates wins and the one that does not loses (Figure~\ref{fig:cross-model-case}). A single pair cannot
attribute the outcome to simulation; Table~\ref{tab:case-behavior-comparisons} lists the selected comparisons.

% Generated by scripts/analyze_case_behaviors.py; do not edit manually.
\begin{table}[t]
\centering
\caption{\textbf{Same-case trajectory contrasts selected by fixed rules.} C1, largest same-case success spread between models; C2, same model and first move, different outcomes; C3, Restricted-observation invalid-action losses versus a peer; C4, explicit first-turn tool use versus none; C5, rejected moves recovered within the turn versus none rejected; C6, different first-moved piece, different outcomes. Trial $k$ is the $k$-th of the three trials in a model--case--setting cell. Root move is the stored reference first move; Played lists up to four agent moves. B/L count successful/attempted explicit board and legal-move calls; S counts accepted/attempted simulation calls, not fully legal lines. Terminal reasons follow the rules replay of Appendix~\ref{app:rules-check}. These purposefully selected examples do not estimate representative effects or internal reasoning.}
\label{tab:case-behavior-comparisons}
\fontsize{9}{11}\selectfont
\setlength{\tabcolsep}{4pt}
\begin{tabular}{@{}p{0.27\columnwidth}p{0.35\columnwidth}p{0.30\columnwidth}@{}}
\toprule
Contrast; model; setting & Root move; agent moves & Outcome; actions \\
\midrule
C1a JH-084\newline Gemini 3.1 Pro, trial 1\newline Sighted & Root move: g5g9.\newline Played: g5g9, h5h9 & Pass; checkmate.\newline Invalid: 0. B/L/S: 0/0, 0/0, 0/0. \\[2pt]
C1b JH-084\newline GPT-OSS 120B, trial 1\newline Sighted & Root move: g5g9.\newline Played: g5d5, e0d0, d0e0, e0d0, ... & Fail; resignation.\newline Invalid: 3. B/L/S: 1/1, 1/1, 0/0. \\[2pt]
\midrule
C2a SQYQ-066\newline Gemini 3.1 Pro, trial 2\newline Sighted & Root move: g8g9.\newline Played: g8g9, d8e8, g9g8, i5g6, ... & Pass; checkmate.\newline Invalid: 0. B/L/S: 0/0, 0/0, 4/7. \\[2pt]
C2b SQYQ-066\newline Gemini 3.1 Pro, trial 1\newline Sighted & Root move: g8g9.\newline Played: g8g9, d8d9, g9g8, i5g6, ... & Fail; attacker checkmated.\newline Invalid: 0. B/L/S: 0/0, 0/0, 8/18. \\[2pt]
\midrule
C3a SQYQ-062\newline DeepSeek R1, trial 2\newline Restricted observation & Root move: f8e8.\newline Played: a5e5, e5e8, b8c6 & Fail; invalid-action limit.\newline Invalid: 5. B/L/S: 0/0, 0/0, 0/0. \\[2pt]
C3b SQYQ-062\newline Gemini 3.1 Pro, trial 1\newline Restricted observation & Root move: f8e8.\newline Played: c8c9, a5e5, e2g0 & Fail; attacker checkmated.\newline Invalid: 1. B/L/S: 0/0, 0/0, 0/0. \\[2pt]
\midrule
C4a JH-327\newline DeepSeek V4 Pro, trial 3\newline Sighted & Root move: h2h9.\newline Played: h2h9, i2d2, d2d4, d4d5 & Pass; checkmate.\newline Invalid: 0. B/L/S: 1/1, 1/1, 2/3. \\[2pt]
C4b JH-327\newline Kimi K2.5, trial 2\newline Sighted & Root move: h2h9.\newline Played: h2h9, i2i9 & Fail; attacker checkmated.\newline Invalid: 0. B/L/S: 0/0, 0/0, 0/0. \\[2pt]
\midrule
C5a SQYQ-098\newline Qwen 3.7 Max, trial 2\newline Restricted observation & Root move: b2b9.\newline Played: b2b9, g7g8, d0d1, d1d2, ... & Pass; defender perpetual check.\newline Invalid: 5. B/L/S: 0/0, 0/0, 0/0. \\[2pt]
C5b SQYQ-098\newline Gemini 3.1 Pro, trial 2\newline Restricted observation & Root move: b2b9.\newline Played: b2b9, h6h9, g7g8, h9f9, ... & Pass; checkmate.\newline Invalid: 0. B/L/S: 0/0, 0/0, 0/0. \\[2pt]
\midrule
C6a SQYQ-012\newline Gemini 3.1 Pro, trial 1\newline Sighted & Root move: g7e7.\newline Played: g7e7, d8e8, d7b8 & Pass; checkmate.\newline Invalid: 0. B/L/S: 0/0, 0/0, 1/1. \\[2pt]
C6b SQYQ-012\newline Gemini 3.1 Pro, trial 2\newline Sighted & Root move: g7e7.\newline Played: d8e8, d7c9, g8f8, f8e8, ... & Fail; resignation.\newline Invalid: 1. B/L/S: 0/0, 0/0, 6/10. \\[2pt]
\bottomrule
\end{tabular}
\end{table}

\end{document}